\documentclass[acmtog,nonacm]{acmart}

\usepackage{booktabs} 
\usepackage{amsmath}
\usepackage{pifont}

\usepackage{enumitem}
\usepackage{subcaption}
\usepackage{graphicx}
\usepackage{paralist}
\usepackage{mathtools}
\usepackage{cleveref}
\usepackage{multirow}
\usepackage{makecell}
\newcommand{\pose}{\mathbf{P}}
\renewcommand{\position}{\mathbf{M}}

\usepackage{xcolor}
\usepackage{float}
\floatstyle{plaintop}
\restylefloat{table}

\setcopyright{none}

\newcommand{\Ex}[1]{\mathop{\mathbb{E}}_{#1}}

\begin{document}

\title{Bringing Diversity from Diffusion Models to Semantic-Guided Face Asset Generation}

\author{Yunxuan Cai}
\orcid{0000-0002-6063-5513}
\authornote{Equal contribution}
\authornote{This is the author version of a work published in \emph{ACM Transactions on Graphics (TOG)}, 2026. DOI: 10.1145/3793859.}
\affiliation{%
  \institution{University of Southern California, USC Institute for Creative Technologies}
  \city{Los Angeles}
  \country{USA}}
\email{ycai@ict.usc.edu}
\author{Sitao Xiang}
\authornotemark[1]
\orcid{0000-0001-9296-6889}
\affiliation{%
  \institution{University of Southern California, USC Institute for Creative Technologies}
  \city{Los Angeles}
  \country{USA}}
\email{sitaoxia@usc.edu}
\author{Zongjian Li}
\orcid{0009-0007-2069-4096}
\affiliation{%
  \institution{University of Southern California, USC Institute for Creative Technologies}
  \city{Los Angeles}
  \country{USA}}
\email{zongjian@usc.edu}
\author{Haiwei Chen}
\orcid{0000-0002-4918-0917}
\affiliation{%
  \institution{USC Institute for Creative Technologies}
  \city{Los Angeles}
  \country{USA}}
\email{chw9308@hotmail.com}
\author{Yajie Zhao}
\orcid{0000-0001-9929-6187}
\affiliation{%
  \institution{USC Institute for Creative Technologies}
  \city{Los Angeles}
  \country{USA}}
\email{zhao@ict.usc.edu}

\begin{abstract}
High-quality 3D face asset creation remains costly due to reliance on controlled capture setups and manual processing, limiting scalability and diversity. We introduce a fully automated, semantically controllable framework for generating PBR-ready 3D facial assets without requiring dedicated scans. Our pipeline begins with a diffusion-based data synthesis stage, where 2D portrait samples from a pre-trained diffusion model are converted into 44K textured 3D face reconstructions via our proposed geometry recovery and texture normalization algorithm, which aligns arbitrarily shaded outputs into clean albedo space. Using this dataset, we train a disentangled adversarial generator that maps semantic attributes (age, gender, ethnicity) to UV-space geometry and albedo, enabling both direct sampling and continuous latent editing while preserving identity. A refinement stage further produces PBR materials and secondary assets (eyeballs, teeth, gums). The resulting system supports controllable face generation and post-editing in real time and exports directly to standard rendering and animation pipelines. We evaluate each component extensively and provide a web-based interactive interface to showcase practical deployment. 
\end{abstract}



\begin{teaserfigure}
  \includegraphics[width=\textwidth]{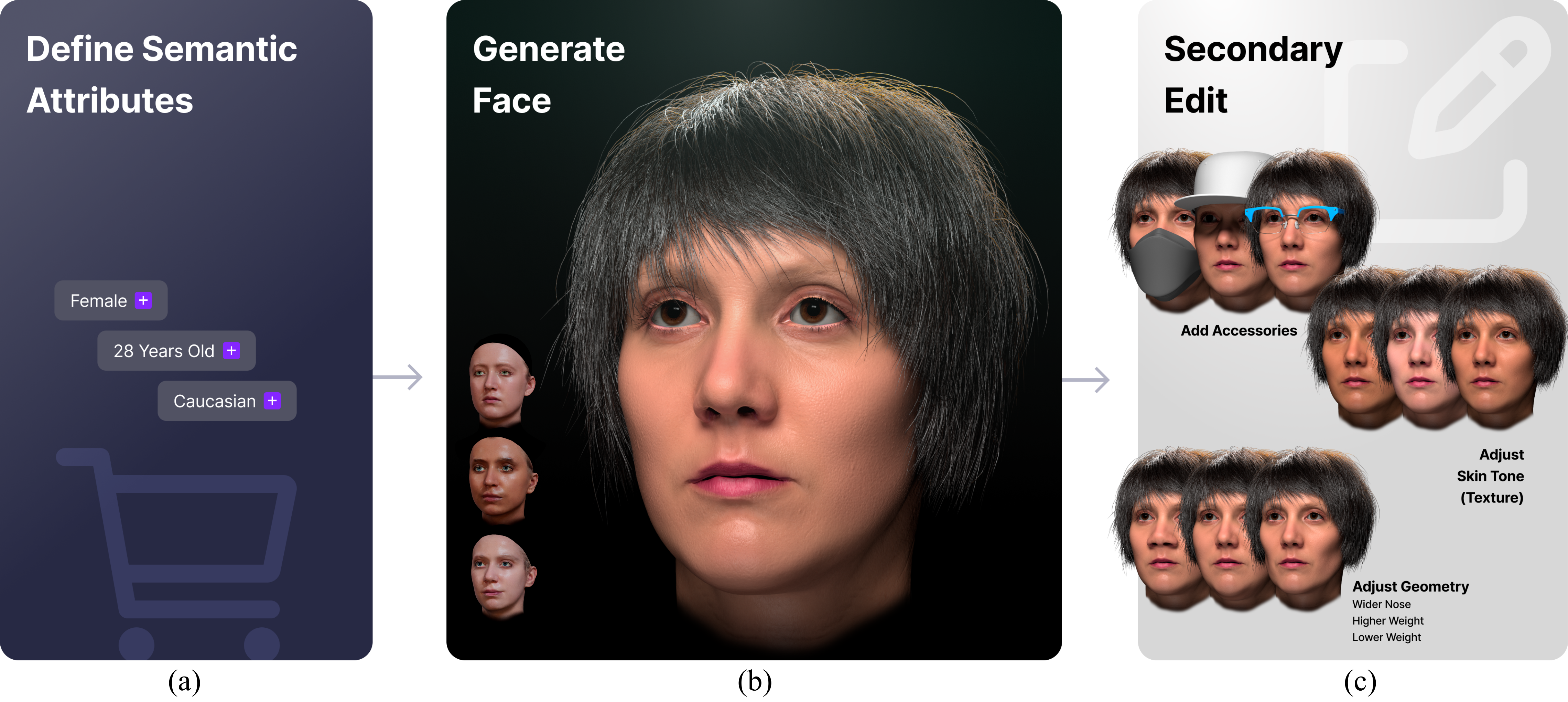}
  \caption{We propose a high-quality, novel semantically controllable 3D face asset generator. This allows users to create customized avatars with a full spectrum of assets for realistic rendering. (a) User-defined semantic labels. (b) Avatars generated and rendered using all the assets. Users can generate multiple avatars and select their preferred subject. (c) Our model also supports texture space and geometry space editing, and offers a handcrafted accessory database (e.g., hairstyles, hats, glasses) for users to choose from.}
  \Description{}
  \label{fig:teaser}
\end{teaserfigure}

\maketitle

\section{Introduction}

Creating realistic 3D human faces has long been a central goal in both industry and academia. A high-quality 3D face asset integrates detailed skin textures, accurate geometry, and material properties to function within a physically based rendering (PBR) pipeline. Demand for such assets continues to surge across gaming, film, teleconferencing, and AR/VR applications. However, producing them remains labor-intensive and expensive—typically requiring skilled artists and, in the case of scans, access to a carefully calibrated facial capture studio. Even today, generating production-ready 3D face assets is still a costly and time-consuming endeavor.

Despite extensive efforts to automate high-quality 3D face creation for broader accessibility, current face generation models remain constrained by the scarcity and imbalance of available facial asset datasets. Industrial systems like MetaHuman~\cite{fang2021metahuman} rely on limited preset assets, while traditional 3D Morphable Models~\cite{DBLP:conf/avss/PaysanKARV09, DBLP:journals/ijcv/BoothRPDZ18, DBLP:journals/tvcg/CaoWZTZ14,DBLP:journals/tog/LiBBL017} and learning-based approaches~\cite{li2020learning, yang2020facescape, zhang2023dreamface} struggle to jointly model geometry and texture, failing to capture the full diversity of real human faces or offer precise semantic control. Data limitations are especially evident in facial texture generation—skin tone, freckles, wrinkles, and pores are poorly represented, leading to biased outputs such as DreamFace~\cite{zhang2023dreamface}, which predominantly synthesizes Asian-like textures due to dataset skew (over 90\% of FaceScape~\cite{yang2020facescape} is Asian). Recent methods that leverage CLIP~\cite{radford2021learning} or large diffusion models improve diversity through knowledge distillation, as seen in AvatarClip~\cite{hong2022avatarclip} and DreamAvatar ~\cite{cao2023dreamavatar}, while LucidDreamer~\cite{EnVision2023luciddreamer} enhances detail using Interval Score Matching. However, these approaches still suffer from artifacts, incomplete assets, slow inference, and lack of animation-ready outputs, leaving a significant gap between generative quality and production usability.

Our approach addresses data scarcity in 3D face asset generation by leveraging large diffusion models for diverse synthesis while introducing three key innovations that bridge the gap between scanned-quality assets and synthetic outputs. (1) Diffusion-generated UV textures are often incomplete and inconsistently shaded, making them unsuitable for PBR rendering. We introduce a texture completion and normalization pipeline to produce clean, fully albedo-compatible maps. (2) We enable precise semantic control and editing by structuring attributes around demographic labels (ethnicity, gender, age) and adopting a disentangled GAN-based framework, which preserves identity during editing more effectively than diffusion-based inversion. (3) We drastically reduce generation time by distilling knowledge into a GAN, achieving 0.014s versus 40s per face compared to a diffusion model on a single A6000 GPU—making high-quality face asset generation both controllable and real-time.

Our training pipeline consists of two major stages. \textit{The first stage} leverages an image-based diffusion network and a 3D face reconstruction network to create a dataset of diverse, high-quality 3D face assets. As the direct image output of the diffusion network contains lighting and other external visual effects, albedo maps are computed from a proposed texture normalization algorithm. In addition, we apply a sanity check to ensure quality and discard any data with artifacts or mismatched labels, and convert the conditioning text used to generate the data into three controlling attributes: age, gender, and ethnicity. This process resulted in a dataset of 44K 3D face models. \textit{The second stage} extends from DisUnknown~\cite{xiang2021disunknown} to train a GAN with the processed training data from the first stage. As described, this stage produces the final model that generates 3D face assets consisting of 4K geometry, albedo, specular and displacement maps in the UV space. Thanks to our unique design on adversarial training, the flexible model can either create 3D assets from an attribute description (``generate a 40-year-old, Hispanic male''), or perform inversion on a given image and subsequently edit and reconstruct a 3D asset from the image, based on an attribute description (``turn a face photo into a 3D face of a 40-year-old, Hispanic male, without losing their identity''). 

We summarize our contributions as follows:

\begin{compactitem}
\item We've introduced a comprehensive, practical and novel framework for generating high-quality face assets. This system uses user-defined semantics and attributes to create PBR-based face assets, including base geometry, albedo, specular, and displacement maps, as well as secondary assets such as eyeballs, teeth, and gums. The system also allows for post-generation editing in both geometry and texture, while preserving identity. The generated face avatars can be seamlessly integrated into downstream applications for rendering and animation. Additionally, we've developed an interactive web UI for users to explore these features.
\item We have developed a large, high-quality 3D face database containing $44K$ albedo/geometry pairs, complete with age, gender, and ethnicity labels. This database exemplifies an effective way to use a pre-trained diffusion model in an industry production pipeline.
\item We also tackled the challenging problem of domain transfer with unbalanced data amounts in two domains. Our texture normalization framework transfers 44,000 unconstrained images into a domain containing 200 images. This allows us to combine the diversity of one domain with the quality from another. This could inspire research in this field.

\end{compactitem}
\section{Related Work}
\subsection{3D Face Morphable Model}

The 3D Morphable Model (3DMM), as the core component of conventional 3D face generation, was first introduced in ~\cite{DBLP:conf/siggraph/BlanzV99} as a compact representation of face models in parametric space. Since then, it has been extensively applied in face recognition~\cite{DBLP:conf/avss/PaysanKARV09}, face reconstruction~\cite{DBLP:conf/accv/BasSBW16,DBLP:conf/cvpr/ThiesZSTN16,gecer2019ganfit}, and avatar creation~\cite{nagano2018pagan,li2020dynamic,DBLP:conf/graphicsinterface/GhafourzadehRFB20}. A comprehensive overview of 3DMM is provided in ~\cite{DBLP:journals/tog/EggerSTWZBBBKRT20}. Efforts have been made to improve the expressiveness, quality, and parameterization capability of 3DMM over the past 20 years. 

Previous work ~\cite{DBLP:conf/avss/PaysanKARV09,DBLP:journals/tvcg/CaoWZTZ14,DBLP:conf/cvpr/BoothRZPD16,DBLP:journals/tog/LiBBL017} reduced the cost and labor required for data acquisition and registration. ~\cite{DBLP:conf/avss/PaysanKARV09} introduced the first publicly available 3DMM, while ~\cite{DBLP:journals/ijcv/BoothRPDZ18} developed a more diverse linear model using approximately 10,000 scans. For expression modeling, blendshapes were introduced to a bilinear model~\cite{DBLP:journals/tvcg/CaoWZTZ14,DBLP:journals/tog/LiBBL017}. Deep learning methods have since enhanced 3DMM capabilities, enabling more diverse and robust representations~\cite{DBLP:conf/wacv/AbrevayaWB18,DBLP:conf/eccv/RanjanBSB18,DBLP:journals/ijcv/DaiPSD20,DBLP:conf/cvpr/LiBZCIXRPKXL20,DBLP:conf/cvpr/SmithSDTTE20,DBLP:conf/3dim/ChandranBGB20}. ~\cite{yang2020facescape,DBLP:journals/tog/LiKZHB020} constructed high-quality 3DMM with pore-level resolution data, and ~\cite{yang2020facescape} provided a large-scale textured 3D face dataset that represents geometry through both rough shapes and detailed displacement maps. ~\cite{DBLP:journals/tog/LiKZHB020} developed a non-linear 3DMM using high-resolution face scans, incorporating material attributes for physically-based rendering. However, challenges remain in expensive data collection and limited diversity.

When applied to 3D face generation, conventional 3DMMs such as ~\cite{DBLP:conf/avss/PaysanKARV09, DBLP:journals/ijcv/BoothRPDZ18, DBLP:journals/tvcg/CaoWZTZ14,DBLP:journals/tog/LiBBL017} offer parameterized models for face modeling. However, they lack explicit control over individual attributes and struggle to effectively integrate textures with underlying geometry. This limitation restricts their usefulness in applications requiring customized human face generation. While learning-based 3DMMs like ~\cite{li2020learning,yang2020facescape} enable joint modeling of texture and geometry, these models still cannot provide adequate semantic or attribute control during generation.

\subsection{Text-to-3D Face}
Advancements in text-to-3D avatar generation~\cite{michel2022text2mesh,hong2022avatarclip,wu2023high} have leveraged the vision-language model CLIP~\cite{radford2021learning} to map natural language descriptions into latent spaces, enabling the generation of 2D images that are then converted into 3D avatars. While this approach allows attribute and semantic control, it often suffers from texture artifacts and inconsistent multi-view outputs. Beyond avatar generation, text-to-3D content creation in general has rapidly evolved, with recent works~\cite{cao2023dreamavatar,chen2023fantasia3d,lin2023magic3d,metzer2023latent,poole2022dreamfusion} excelling at generating diverse 3D models from text. These methods employ pre-trained text-to-image diffusion models as priors to guide the training of parameterized 3D models, ensuring multi-view consistency and text alignment through Score Distillation Sampling (SDS). However, SDS encourages average-seeking behaviors in the 3D representation, leading to over-smoothed results. LucidDreamer~\cite{EnVision2023luciddreamer} addresses this issue with Interval Score Matching (ISM), improving detail quality, but their method remains computationally expensive — taking approximately 36 minutes to generate a 3D model. 
A challenge in text-to-3D generation is that the appearance often entangles reflectance and illumination. Intrinsic decomposition addresses this issue by separating reflectance (albedo) from lighting, and diffusion priors~\cite{chen2024intrinsicanything,luo2024intrinsicdiffusion,kocsis2024intrinsic,li2025idarb} have recently been used to improve such decomposition in a general image.
At the same time, text-to-3D face avatar generation requires producing a usable 3D facial asset, which imposes additional constraints than those for general images.

Among existing methods, DreamFace~\cite{zhang2023dreamface} has made significant progress in generating high-quality face avatars. It separately trains a geometry and appearance model, fine-tuning a generic diffusion model with 618 textures from sources like FaceScape \cite{yang2020facescape} and 3DScanStore~\cite{3DScanStore2023}. However, several limitations exist:\textbf{1.)} Limited geometric diversity – Geometry is initialized from a 3DMM with only 100 bases. \textbf{2.)} Insufficient and biased texture data – 618 textures are inadequate for fine-tuning a diffusion model trained on vast in-the-wild datasets, and FaceScape's 90\% Asian demographic introduces bias. \textbf{3.)} Decoupled geometry and texture generation – Ignoring their correlation, as highlighted in~\cite{li2020learning}. \textbf{4.)} Challenging inversion in diffusion models – Editing (e.g., tattoos, makeup) is constrained to one-pass modifications without precise control. \textbf{5.)} Lack of quality control – Outputs are inconsistent, and quality depends heavily on user prompts.

Inspired by DreamFusion~\cite{poole2022dreamfusion}, LucidDreamer~\cite{EnVision2023luciddreamer} and HeadStudio~\cite{zhou2024headstudio}, we aim to refine generative 3D generation. While these models generate 3D faces from text, they often produce artifacts and require manual quality control, making them unsuitable for direct application. Nonetheless, pre-trained image diffusion models, trained on large datasets, naturally enhance diversity, provide annotations and offer great accessibility. Our approach harnesses the rich priors of image diffusion models to improve quality, control, and usability in 3D avatar generation.

\subsection{Semantic Face Generation and Manipulation}

Generative Adversarial Networks (GANs), pioneered by \cite{DBLP:conf/nips/GoodfellowPMXWOCB14}, have been well studied in the setting of semantic editing. A key challenge lies in achieving semantic and disentangled control in generative models, \textit{e.g.}, randomly changing one specific attribute while preserving the other attributes. Efforts to interpolate in the latent space for smooth output variation have been explored by ~\cite{DBLP:conf/iclr/Laine18,DBLP:conf/cvpr/Shao0F18}, while recent research focuses on disentangling the latent space for semantic control~\cite{DBLP:conf/cvpr/ShenGTZ20,DBLP:conf/nips/HarkonenHLP20,DBLP:journals/corr/abs-2007-06600,DBLP:conf/iccv/0003HPL021,DBLP:conf/cvpr/ZhengHTSS21}. Observations of vector arithmetic in latent space by ~\cite{DBLP:journals/corr/RadfordMC15,DBLP:conf/cvpr/UpchurchGPPSBW17} have led to unsupervised disentanglement methods. Semi-supervised methods apply principal component analysis for attribute identification~\cite{DBLP:conf/nips/HarkonenHLP20}, while supervised methods use labeled data for latent space factorization~\cite{DBLP:conf/cvpr/ShenGTZ20,DBLP:conf/eccv/KowalskiGEBJS20}. Alternative approaches utilize 3DMM for semantic control over pre-trained StyleGAN latent space~\cite{DBLP:conf/cvpr/TewariEBBSPZT20, DBLP:journals/corr/abs-2104-11228} or train unsupervised data with labels from attribute classifiers~\cite{DBLP:conf/wacv/KhodadadehGMLBK22}. Also, to transfer such controllability to real image editing, the GAN inversion methods~\cite{DBLP:conf/iccv/AbdalQW19,DBLP:conf/eccv/ZhuSZZ20,DBLP:journals/tog/TewariE0BSPZT20} propose to map the real image into the latent space and thus can manipulate real images.
Recently,~\cite{fadaeinejad2025geometry} incorporated artist input into a GAN-based framework, enabling greater controllability in the avatar creation workflow.

\subsection{Facial Texture Synthesis}

3D facial texture synthesis has evolved from traditional geometric and photometric methods to advanced neural network-based techniques. Traditional photometric methods, relying on polarized light to capture skin reflectance properties, laid the groundwork for detailed texture mapping~\cite{debevec2000acquiring, ghosh2011multiview}. Innovations then made high-quality, single-shot captures possible for both facial geometry and reflectance~\cite{riviere2020single, lattas2022practical}.

The advent of deep learning has revolutionized texture synthesis. Neural networks are employed for photorealistic textures, combining low and high-frequency methods~\cite{saito2017photorealistic, yamaguchi2018high, chen2019photo, huynh2018mesoscopic}. Generative adversarial networks (GANs) have further enhanced the generation of detailed and realistic textures, using patch-based approaches and focusing on physical texture properties like diffuse and specular albedos~\cite{gecer2019ganfit, lattas2020avatarme, lattas2021avatarme++, dib2021towards, lattas2023fitme}. Recent developments employ the denoising diffusion probabilistic models and text inputs for texture creation~\cite{zhang2023dreamface}. Super-resolution techniques enable the generation of high-resolution textures from lower-resolution inputs, enhancing detail of mesh without altering its geometry~\cite{chen2023dual}.

These advancements underscore the goal towards both visual realism and adherence to the principles of physical-based rendering, which guarantees accurate material properties under diverse lighting conditions.
\section{Method}


Our goal is to create a 3D avatar synthesis model that uses facial attributes (ethnicity, gender and age) to produce high-quality 3D face assets, including 3D geometry and PBR-based material properties such as albedo, specular and displacement maps. To achieve our goal, we start by constructing a large, high-quality and diverse 3D face assets dataset associated with corresponding annotations using a pre-trained diffusion model. We introduce a unique framework in \Cref{sec:data_prep} to construct a 3D face asset dataset that is clean and complete with corresponding labels. Then, a dedicated generative network, as detailed in \Cref{sec:generator}, is trained using the obtained dataset to create the basic geometries and albedo textures. Lastly, we apply post-processing, as described in \Cref{sec:postprocessing}, to refine the generated face geometries and albedo textures and generate the specular and displacement maps to complete the PBR assets.

\begin{figure*}[t!]
\centering
\includegraphics[width=0.9\linewidth]{images/data_prep_v8.pdf}
\caption{Overview of the proposed dataset preparation method: Portraits are initially synthesized using a latent diffusion model that is conditioned by semantic facial attributes and frontal view normal maps. A pre-trained face reconstruction model is then applied to these portraits to extract modified geometries in the form of position maps. Textures are completed by blending the projected portraits with physically-based rendered texture maps from the scanning database. }
\Description{}
\label{fig:data_prep_fig}
\end{figure*}

\subsection{Data Preparation}
\label{sec:data_prep}

\begin{figure}
\centering
\includegraphics[width=\linewidth]{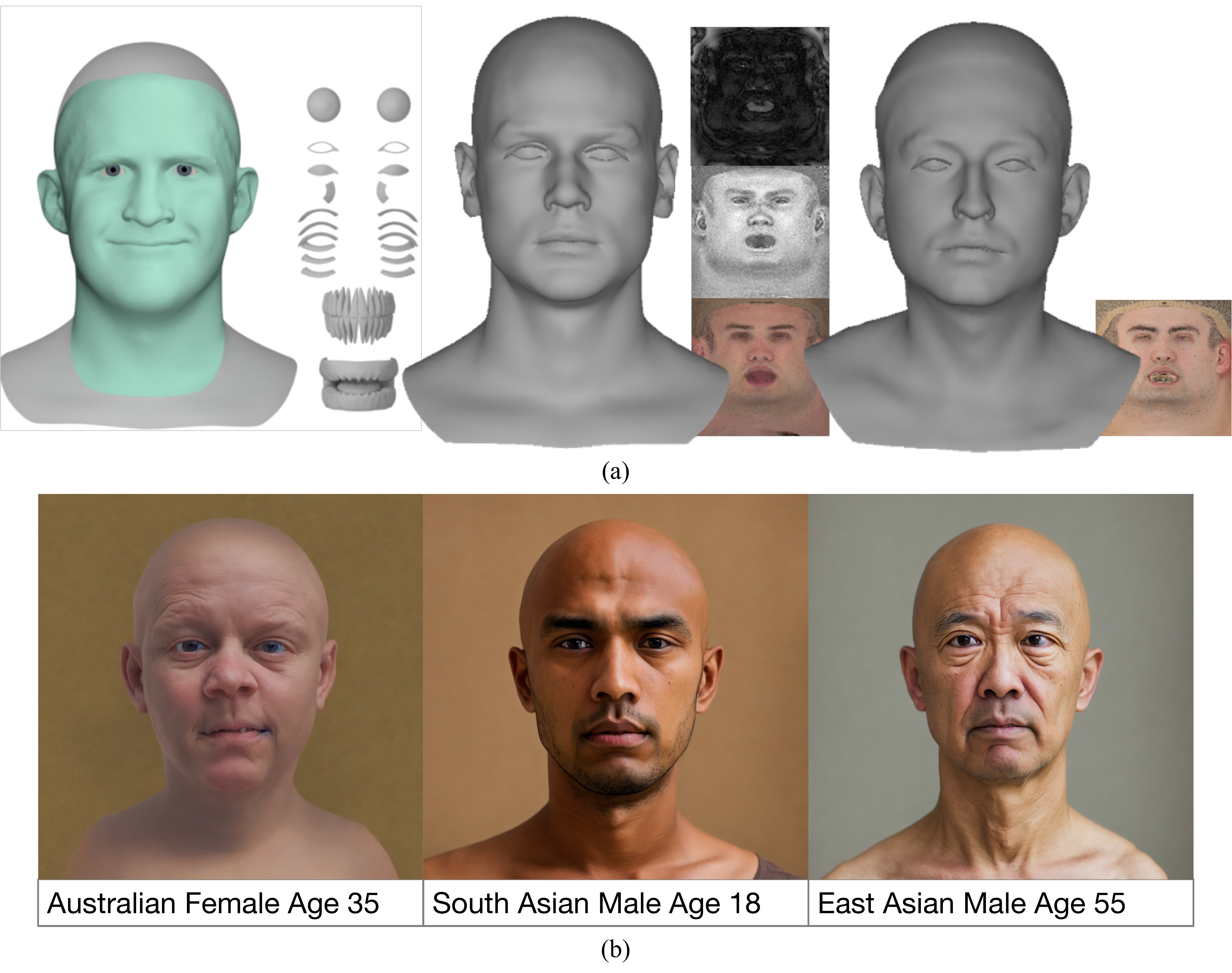}
\caption{Two types of 3D human face data. (a) From left to right: the template model used for registering all data resources, a sample of Light Stage data, and a sample of Triplegangers data. (b) Samples of semantic attributes and the resulting 2D portraits from LDM.}
\Description{}
\label{fig:dataillustration}
\end{figure}

\begin{figure}[th!]
\centering
\includegraphics[width=1.0\linewidth]{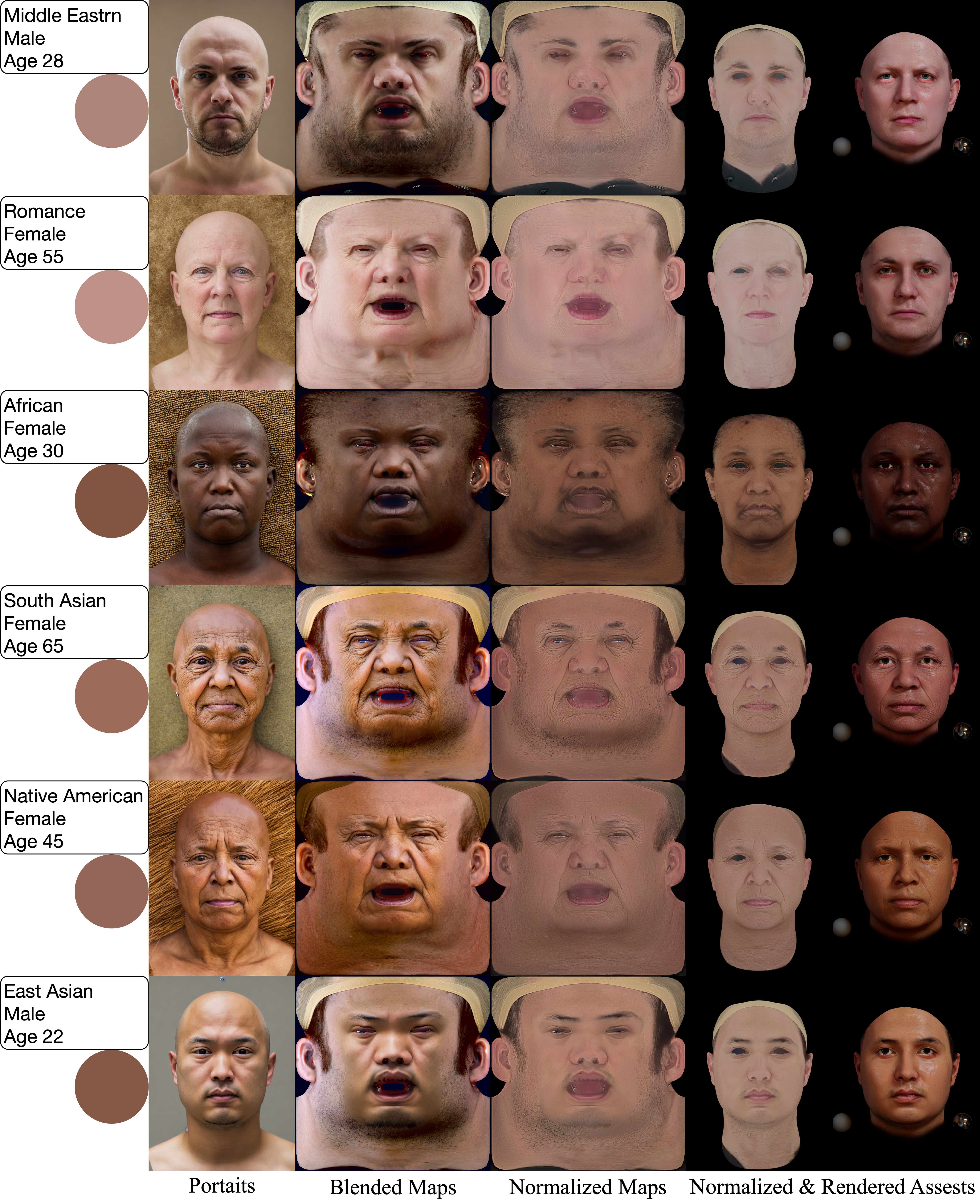}
\caption{Examples of training data. From left to right in each example are: the input attribute (semantic and skin tone guide), portrait, map before normalization, normalized texture map, and images rendered with and without post-processing. (\Cref{sec:normalization}).}
\Description{}
\label{fig:normarlizationresults}
\end{figure}


We leverage the rich priors of an image diffusion model by using it to synthesize a dataset of high-quality 3D faces with annotations, which is used to train our Text2Avatar 3D face asset generator. This section details three main steps needed to create a dataset $\mathcal{D}_{main}$ of attribute-controlled 3D assets (shapes and albedo textures). Section~\ref{sec:port_gen} describes the generation of 2D portraits that leverages information from both the rich priors of a large image diffusion model~\cite{rombach2022high, zhang2023adding} and a dataset $\mathcal{D}_{scan}$ of scanned 3D faces from LightStage~\cite{li2020learning} and the TrippleGangers dataset~\cite{triplegangers}. Section~\ref{sec:text_comp} describes the construction of 3D face geometries and complete UV textures from the generated portraits. Section~\ref{sec:normalization} details a normalization method that removes lighting and other external visual effects from the textures to create the clean albedo maps.

\subsubsection{Portraits Generation using Diffusion Models}
\label{sec:port_gen}

We utilize the pre-trained text-to-image Stable Diffusion v1.5~\cite{rombach2022high} and ControlNet~\cite{zhang2023adding} (trained on normal maps) to generate a rich collection of 2D portraits. Specifically, the diffusion model is given various text descriptions of facial attributes. To avoid ambiguity caused by the language model and ensure precise control, we define three major demographic attributes that contribute most to human appearances and shapes: ethnicity, gender, and age. We also include generic prompts for illumination and quality control, like resolution and framing. For example, "East Asian female age 20" is one set of demographic attribute combinations. Age groups span from 15 to 75 years old. For gender, we use the labels male, female, and unisex (a gender-neutral semantic category positioned between male and female). We also consider ethnicity/race and geographic location, resulting in 14 distinct categories: \textit{African, East Asian, Southeast Asian, South Asian, Middle Eastern, Caucasian, Germanic, Celtic, Slavic, Romance, Australian, Native Americans, Aboriginal, and Pacific Islander.} In addition, we condition the portrait generation with normal maps, rendered from a randomly sampled face geometry at the frontal view, from the scan dataset $\mathcal{D}_{scan}$. The normal maps provide head pose and geometry guidance to constrain the synthesized image. 

Since random sampling may create a conflict between the desired attributes and the selected geometry, we further reconstruct a refined face geometry based on the alignment between the generated portrait and the sampled face geometry using ReFA~\cite{ReFA} (see Section~\ref{sec:imp_details} for details). 

In total, 65K portraits with 3D geometry are created at this stage. Ethnicity is uniformly sampled from 14 categories. As for gender, the distribution consists of $45\%$ male, $45\%$ female, and $10\%$ unisex. Age is uniformly sampled from the following groups: $15$, $18$, $20$, $22$, $25$, $28$, $30$, $35$, $40$, $45$, $55$, $65$, and $75$.


\subsubsection{Texture Completion}
\label{sec:text_comp}

The first step produces a collection of 2D portraits, each with the corresponding attribute labels and predicted geometry. To process this data into 3D assets, we complete the UV-space texture maps from these outputs. Specifically, the texture map is initialized by directly projecting the 2D portraits based on the predicted geometry. However, since only the frontal part of the face is visible, we propose a completion method to fill in the missing regions (e.g. ears) in the boundary of the UV maps. We first collect a dataset of skin textures $\mathcal{D}_{texture}$ by baking randomized lighting into the UV-space textures of the scan dataset $\mathcal{D}_{scan}$ (see \Cref{fig:data_prep_fig}). Then, given the projected partial textures, we search for the nearest neighbor with the highest similarity to the projected portrait in $\mathcal{D}_{texture}$, using Peak Signal-to-Noise Ratio (PSNR) as our measure. The complete UV-space texture is then obtained by blending the projected partial textures and the retrieved texture maps with the pyramid blending algorithm~\cite{burt1983multiresolution}.  

Finally, to ensure the quality of the completed textures, a sanity check is applied to filter out results with artifacts to maintain high data quality. The filtering step keeps approximately $44K$ UV textures from the original $65K$ portraits. However, these textures still contain baked-in lighting, which is not part of our desired clean albedo. Therefore, a further normalization step is proposed in the next section.

\subsubsection{Texture Data Normalization}
\label{sec:normalization}

The goal of this step is to remove lighting and other external visual effects, such as makeup, facial hair, glasses, and shadows, from the UV textures created in the last step. The normalized texture can be considered a clean albedo map that is a part of the output 3D face asset. FFHQ-UV~\cite{bai2023ffhq} is the latest work that can normalize face texture into an albedo image, but we observe that it still struggles to fully preserve identities and facial attributes when applied to our synthesized portraits (see Table~\ref{tab:bs_error}). We therefore propose a more effective normalization method, by posing texture normalization in this context as a domain transfer problem from the source domain, which consists of synthetic textures under different lighting as shown in~\Cref{fig:dataillustration} (b), to the target domain, which consists of clean albedo as shown in~\Cref{fig:dataillustration} (a). Since the target domain contains only 200 identities while the source domain contains around 44K identities, a convolutional image-to-image translation model directly trained on the whole image would face a high risk of overfitting. The following text describes several designs to mitigate this issue.

First, we assume that lighting affects the appearance of the face approximately uniformly in a reasonably-sized patch. Therefore we divide an input image into a $64\times64$ grid of patches and
assign a spatially varying factor $\theta$ to each grid corner to account for factors that are independent of albedo. $\theta$ is bilinearly interpolated in the interior of each patch. Under this formulation, image translation is formulated as a function $f(r, g, b; \theta)$ parameterized by an MLP network that takes the image as input and translates each pixel independently based on the spatially-varying factor $\theta$. However, the image translation is preceded by a patch parameter estimation network, a convolutional network that takes the image as input and computes $\theta$ at patch corners. Effects such as subsurface scattering and inaccuracies due to the non-physical nature of the input images can also be handled by the spatial variation of $\theta$ and do not need to be considered separately.

\begin{figure}
\centering
\includegraphics[width=\linewidth]{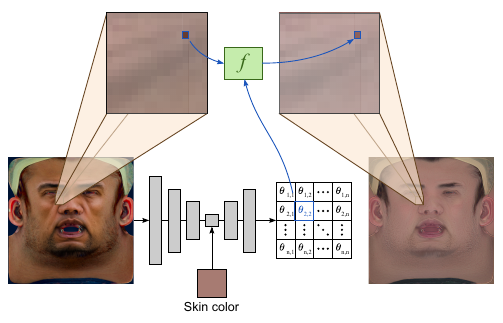}
\caption{Structure of the normalization model. It consists of two networks: the patch parameter estimation network and the pixel translation network.}
\Description{}
\label{fig:norm_net}
\end{figure}

In addition, different combinations of albedo and lighting can result in the same observed color, so without knowing the true lighting, the albedo is ambiguous. To remove this ambiguity, we compute a skin color by taking the average pixel value over manually selected flat regions of the input face (cheeks and forehead), which is equivalent to a masked average of the image $C(x)$. This value is then provided to the patch parameter estimation network explicitly. Together, the patch parameter estimation network and the MLP translation network are termed the normalization model $N(x, C(x))$, which is visualized in \Cref{fig:norm_net}. 

The loss function for training the texture normalization model is constructed as
$
    \mathcal{L}_{\text{N-rec}}=\Ex{x\in X_{\text{scan}}}\big[\|N(x,C(x))-x\|_2\big],
$
where $X_{\text{scan}}$ is the set of scanned albedo textures. During training, we additionally augment the input with albedo images that do not contain lighting and optimize the network to produce identical output to the input in this situation. 

\begin{figure}
\centering
\includegraphics[width=\linewidth]{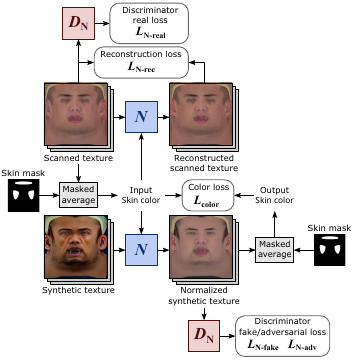}
\caption{Training pipeline for the normalization model. Colored boxes represent learnable models. Identical boxes are the same network (i.e. same weights). Red annotations represent loss functions.}
\Description{}
\label{fig:norm_train}
\end{figure}

The overall training procedure is shown in \Cref{fig:norm_train}. As commonly done in unpaired image-to-image translation, a patch-based discriminator is employed to ensure that the translated image is in the target domain, trained with the binary cross entropy loss. For adversarial training, the distribution of skin color of the normalized synthesized textures is optimized to match that of the scanned textures, so when normalizing a synthesized texture, the skin color is drawn randomly from the set of skin colors of the scanned textures. Let $D_\text{N}$ be the discriminator and $X_{\text{syn}}$ the set of synthetic textures. The Discriminator's loss and the normalization model's adversarial loss are:

\begin{align}
    \mathcal{L}_{\text{N-real}}&=\Ex{x\in X_{\text{scan}}} \big[-\ln D_\text{N}(x)\big],\\
    \mathcal{L}_{\text{N-fake}}&=\Ex{\substack{x_1\in X_{\text{scan}}\\x_2\in X_{\text{syn}}}} \big[-\ln(1-D_\text{N}(N(x_2,C(x_1))))\big],\\
    \mathcal{L}_{\text{N-adv}}&=\Ex{\substack{x_1\in X_{\text{scan}}\\x_2\in X_{\text{syn}}}} \big[-\ln D_\text{N}(N(x_2,C(x_1)))\big].
\end{align}

Additionally, we need to ensure that when normalizing a synthesized texture, the output does have the same skin color as the provided skin color:

\begin{equation}
    \mathcal{L}_{\text{color}}=\Ex{\substack{x_1\in X_{\text{scan}}\\x_2\in X_{\text{syn}}}} \big[\|C(N(x_2,C(x_1)))-C(x_1)\|_2\big].
\end{equation}

$N$ and $D_\text{N}$ are then trained using:

\begin{align}
    \min_{D_\text{N}}&\quad \mathcal{L}_{\text{N-real}}+\mathcal{L}_{\text{N-fake}},\\
    \min_{N}&\quad \lambda_{\text{N-rec}}\mathcal{L}_{\text{N-rec}}+\lambda_{\text{color}}\mathcal{L}_{\text{color}}+\lambda_\text{N-adv}\mathcal{L}_\text{N-adv}.
\end{align}

\subsection{Base Geometry and Albedo Generation}
\label{sec:generator}

\begin{figure*}
\centering
\includegraphics[width=\linewidth]{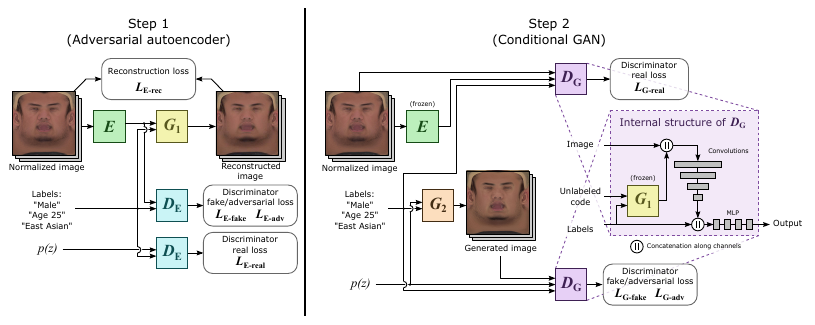}
\caption{2-step training pipeline for the generator. Colored boxes represent learnable models. Identical boxes are the same network (i.e. same weights). Red annotations represent loss functions. $E$ and $G_1$ are frozen in step 2.}
\Description{}
\label{fig:disunknown}
\end{figure*}

As mentioned in the introduction, we aim to learn a geometry and albedo texture generation model that can 1) effectively disentangle labeled attributes and unlabeled information (e.g. identity), 2) provide an easy way to perform image inversion, and 3) generate efficiently. With the prepared dataset (geometry, albedo texture with labeled attributes), we propose a two-step generative model, extending from the two-step approach in~\cite{xiang2021disunknown}, that satisfies all previous requirements, and we show in \Cref{sec:ablation} that the design is more effective compared to the common practices of generation directly conditioning on the given attributes. The overall procedure is shown in \Cref{fig:disunknown}.


\subsubsection{Learning unlabeled information.} In the first step, a GAN with an autoencoder $E$, a generator $G_1$ and a discriminator $D_E$ is designed to disentangle unlabeled information from the labels (in our case, the gender, age and ethnicity attributes). $E$ is first used to extract unlabeled information from the images and discards labeled information. The code computed by the encoder therefore describes information that is independent of the face attributes (e.g. identity information). 

How can $E$ be trained to disentangle information from attribute labels? If it does discard the labeled information completely, then the conditional distribution of the code, on any value of the attributes should be the same. Based on this observation, we can achieve label disentanglement by using an adversarial classifier that classifies the encoder's output by their label, and the encoder $E$ is trained to make the classifier fail. However, while the conditional distribution of the code is learned to be the same given any label, this distribution has no closed form and cannot be easily sampled. We would like to generate new images, so this conditional distribution must be identical to some known, simple prior. So, different from~\cite{xiang2021disunknown}, the code classifier is replaced with a conditional code discriminator.

Given attribute values as conditions, the code discriminator $D_{\text{E}}$ learns to distinguish between the prior distribution and the codes computed by the encoder. We choose the normal distribution as the prior. Let $l(x)$ be the attribute label of a training image $x$, and $X$ be the training dataset, which are pairs of albedo and geometry:

\begin{align}
    \mathcal{L}_{\text{E-real}}&=\Ex{\substack{x\in X\\z\sim p(z)}}\big[-\ln D_\text{E}(z,l(x))\big],\\
    \mathcal{L}_{\text{E-fake}}&=\Ex{x\in X} \big[-\ln(1-D_\text{E}(E(x),l(x)))\big],\\
    \mathcal{L}_{\text{E-adv}}&=\Ex{x\in X} \big[-\ln D_\text{E}(E(x),l(x))\big].
\end{align}

$E$ will also need to retain all unlabeled information. This is achieved using a reconstruction loss. It is worth noting that the first-step generator will not be used as the generator in our finished texture generation model, as it only assists in training the encoder. In addition, $G_1$ takes the labels $l(x)$ as an input, since labeled information is removed from the codes $E(x)$. The overall reconstruction loss is:

\begin{equation}
    \mathcal{L}_{\text{E-rec}}=\Ex{x\in X}\big[\|G_1(E(x),l(x))-x\|_2\big].
\end{equation}

$E$, $G_1$ and $D_\text{E}$ are then trained using:

\begin{align}
    \min_{D_\text{E}}&\quad \mathcal{L}_{\text{E-real}}+\mathcal{L}_{\text{E-fake}},\\
    \min_{E,G_1}&\quad \lambda_{\text{E-rec}}\mathcal{L}_{\text{E-rec}}+\lambda_{\text{E-adv}}\mathcal{L}_{\text{E-adv}}.
\end{align}

\subsubsection{Training the label-conditioned generator.} As mentioned previously, the trained $G_1(E(x),l(x))$ from step 1 does not model the marginal distribution of the attribute labels as it requires unlabeled information. In the second step, a separate conditional generator $G_2(z,l(x))$ is trained to generate base geometry and albedo maps from attributes and a sampled random code $z$. In this step, the trained autoencoder $E$ is frozen. A new discriminator $D_G$ receives the unlabeled code $E(x)$ as a condition and discriminates whether the generated image has preserved the unlabeled information provided by the code. Specifically, a ``real'' sample is a triplet consisting of a training image $x$, its code $E(x)$ and its label $l(x)$, while a ``fake'' sample is a triplet consisting of a random code $z$, random labels $l(x)$ which are produced by taking the labels of a random training image, and $G_2(z,l(x))$, the image generated using the code and the labels:

\begin{align}
    \mathcal{L}_{\text{G-real}}&=\Ex{x\in X}\big[-\ln D_{\text{G}}(x,E(x),l(x))\big],\\
    \mathcal{L}_{\text{G-fake}}&=\Ex{\substack{x\in X\\z\sim p(z)}} \big[-\ln(1-D_{\text{G}}(G_2(z,l(x)),z,l(x)))\big],\\
    \mathcal{L}_{\text{G-adv}}&=\Ex{\substack{x\in X\\z\sim p(z)}} \big[-\ln D_{\text{G}}(G_2(z,l(x)),z,l(x))\big],
\end{align}

and the training procedure of the second step is a plain GAN:

\begin{align}
    \min_{D_{\text{G}}}&\quad \mathcal{L}_{\text{G-real}}+\mathcal{L}_{\text{G-fake}},\\
    \min_{G_2}&\quad \mathcal{L}_{\text{G-adv}.}
\end{align}

Through experiments, we noticed that although concatenating the one-hot attribute labels to the fully connected part of the discriminator worked as intended, incorporating the unlabeled code in the same way did not give satisfactory results. We speculate that the difficulty lies in the observation that the unlabeled code is much longer than the attribute labels and has a strong spatial structure which the discriminator had to learn from scratch. Therefore, we help the discriminator by generating an image from the label and code it receives using the first-step generator $G_1$ and concatenating its output with the input image, so that the spatial structure is provided by the condition image and need not be re-learned. The detailed architecture of our discriminator is shown in the inset in \Cref{fig:disunknown}. $G_1$ remains frozen and is not updated along with other parameters of the discriminator.

\section{Experiments}

\subsection{Implementation Details} 
\label{sec:imp_details}

\paragraph{Synthetic Training Data Generation.}
To generate the portraits~\Cref{fig:data_prep_fig}, we use the pre-trained Stable Diffusion v1.5~\cite{rombach2022high} as the base LDM model and use the ControlNet checkpoint pre-trained on normal maps~\cite{ControlNet2023} to add conditional controls for our frontal view normal map. The sampling method is DDIM with 40 time steps. Classifier-Free Guidance~\cite{ho2022classifier} scale is 9.0. The output portrait resolution from the LDM model is 1024 $\times$ 1024.


\paragraph{Single-View Face Reconstruction.}
In this paper, we perform single-view 3D face reconstruction to predict the 3D geometry of the generated portraits. To train the model, we rendered around $100K$ synthetic frontal view portraits under different illuminations by using assets in our 3D high-quality database. We combined these with captured multi-view real data. A variant of ReFA~\cite{ReFA} was used as the baseline to train the model, where the network is modified to reconstruct from a single-view image. The camera optimization is kept fixed during training, as the ground truth pose $\pose \in \mathbb{R}^{3 \times 4}$ is known. Therefore, in each network step, only the position map $\position$, which corresponds to the 3D geometry of the face, is updated. 

The training of the reconstruction model is performed on NVIDIA A100 graphics cards. The network parameters are randomly initialized and are trained using the Adam optimizer for 120,000 iterations with a learning rate set to $3 \times 10^{-4}$. For the recurrent face geometry optimizer, we set the inference step to $T = 10$, the grid resolution to $r = 3$, the search radius to $c = 1\textrm{mm}$.

\begin{figure}[t]
 \begin{center}
  \centering
  \includegraphics[width=1\linewidth]{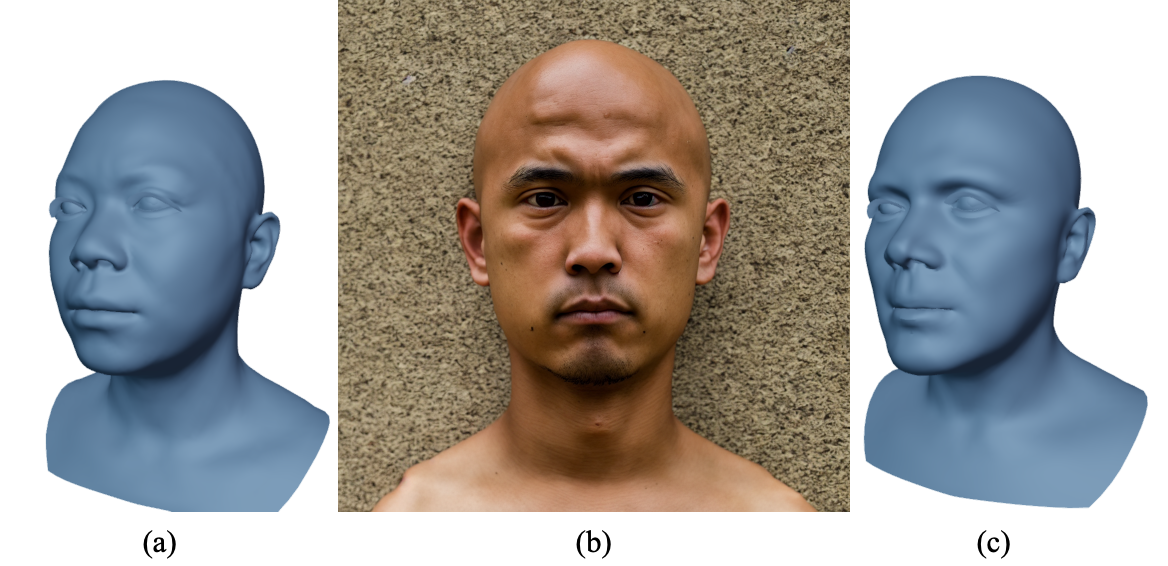}
 \end{center}
 \caption{Results of the single-view face reconstruction model on the portrait, (a) the initial geometry (b) the generated portrait using the initial geometry, (c) the refined geometry with single-view face reconstruction.}
 \label{fig:result_diff}
 \Description{}
\end{figure}

An example of the refinement to the initial geometry is shown in~\Cref{fig:result_diff}, the refined geometry (c) more accurately matches the portrait (b) compared to the initial geometry (a), with deeper-set eyes that are correctly positioned, a mouth that is properly sized and shaped, more correctly aligned cheekbones, and a chin that reflects the correct depth and contour, resulting in an overall head shape that aligns more closely with (b).

\begin{figure}
\centering
\includegraphics[width=0.9\linewidth]{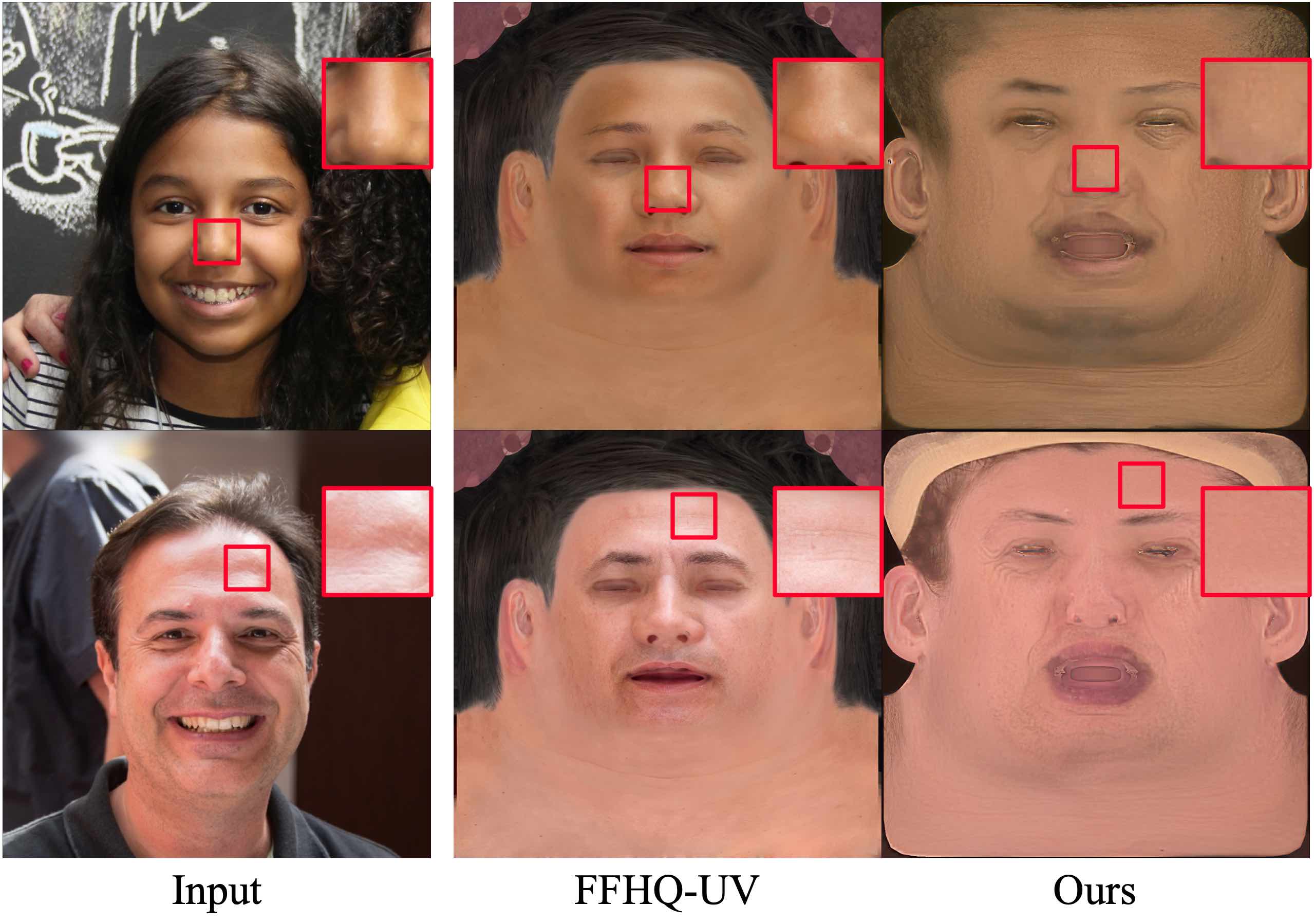}
\caption{Comparison of texture normalization against FFHQ-UV\cite{bai2023ffhq}. No highlights are baked into our results.}
\Description{}
\label{fig:ffhquv}
\end{figure}

\paragraph{Texture Normalization.}
The Patch Parameter Estimation Network is a convolutional network, consisting of alternating kernel 3, stride 1 convolutions and kernel 4, stride 2 convolutions. The number of output channels starts at 16 and doubles after each stride 2 layer. Once the spatial size reduces to 64, the skin color code is broadcast to each location, and two additional kernel 1, stride 1 layers are added. The final layer, which outputs $\theta$, has 8 output channels. The Pixel Translation Network is an MLP with 6 layers and 64 features in the hidden layers.

To ensure that the discriminator focuses on only the lighting, its capacity is purposefully limited: it consists of an MLP with 6 layers and 64 features in the hidden layers. It takes $16 \times 16$ patches as input and first downsamples them to $4 \times 4$ so that the number of input features is $4\times 4\times 3=48$.

\paragraph{Generator}
We adopt the StyleGAN2~\cite{karras2020analyzing} architecture for our generator as well as the convolution part of our discriminator. We also adopt the gradient penalty, path penalty and augmentation from its training procedure.

\paragraph{Asset Refinement}
\label{sec:postprocessing}
We apply an asset refinement process that augments resolution and supplements missing components required for physically-based rendering. This process consists of the following three stages: First, a super-resolution network~\cite{chen2023dual} is used to upsample the 1K albedo map to 4K, recovering high-frequency skin details. Second, a texture translation network~\cite{liang2021high} predicts corresponding specular and displacement maps from the 4K albedo. The network is trained on our high-quality scanned dataset. To generate displacement maps, the albedo is converted to Lab color space and only the L channel is used as input, isolating geometric features from skin color. We also applied a filter that smooths out the edge of the displacement maps. Finally, we integrate secondary assets such as eyeballs, teeth, and gums, along with predefined Blendshapes, into the base mesh to support animation and rendering.

\paragraph{Performance.}
Our base geometry and albedo generation network generates results in 0.014 seconds per subject. For 1K to 4K albedo upsampling, we trained a network~\cite{chen2023dual} on our high-quality dataset, which takes 53 seconds to complete on average. We trained two separate translation networks~\cite{liang2021high} for the specular and displacement maps and the whole translation takes 71 seconds on average. Utilizing these three types of networks, production-quality face assets for an identity can be created in less than 3 minutes on one Nvidia A6000 GPU.

\begin{figure*}
\centering
\includegraphics[width=0.9\linewidth]{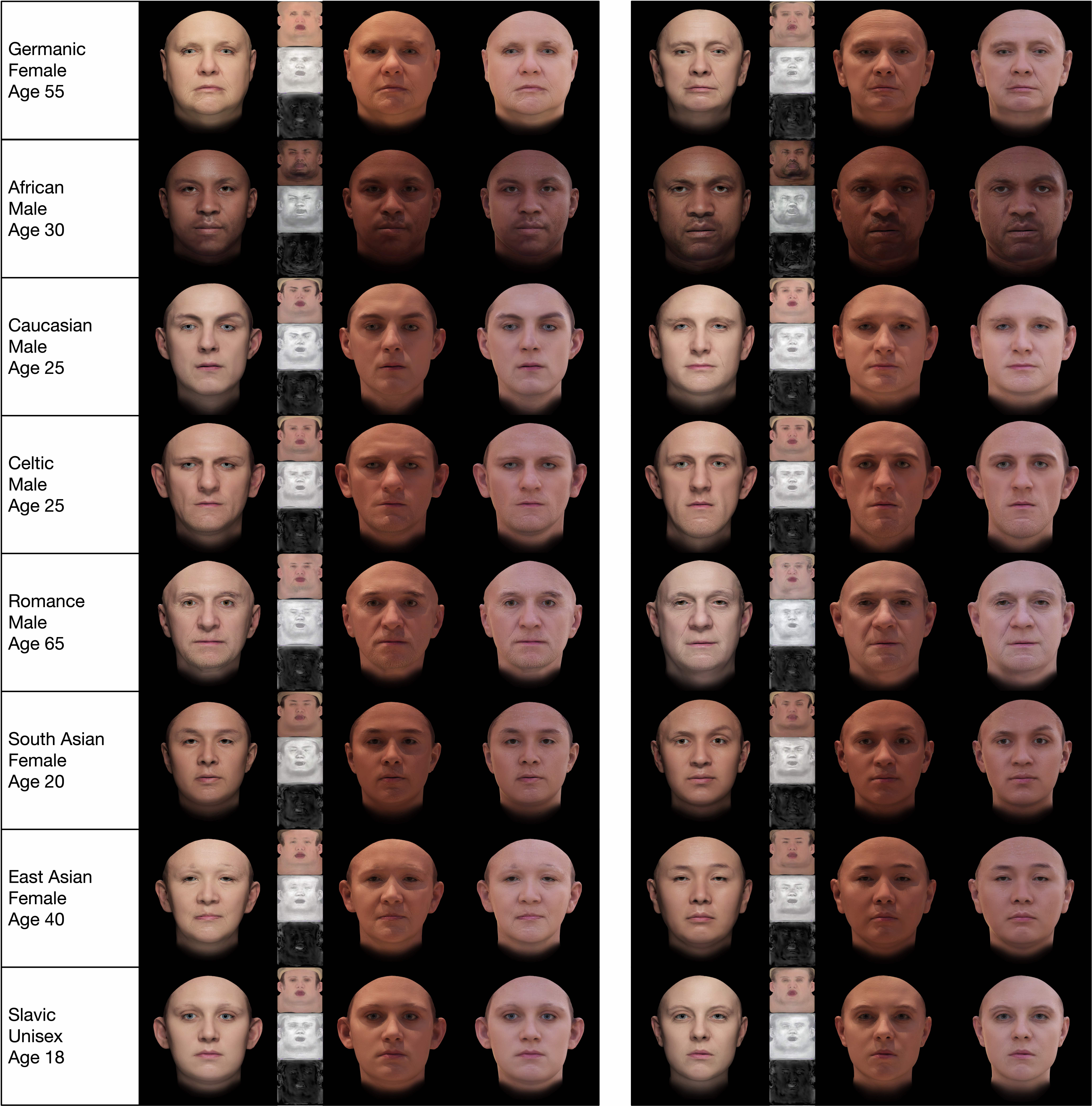}
\caption{Results of generated face models. For the same semantic input in each row, we generate two face models. For each of the generated face models, we show semantic labels, rendering results, PBR assets generated by the method, and rendering results under two other different lighting conditions. }
\Description{}
\label{fig:res_assets}
\end{figure*}

\begin{figure*}
\centering
\includegraphics[width=0.95\linewidth]{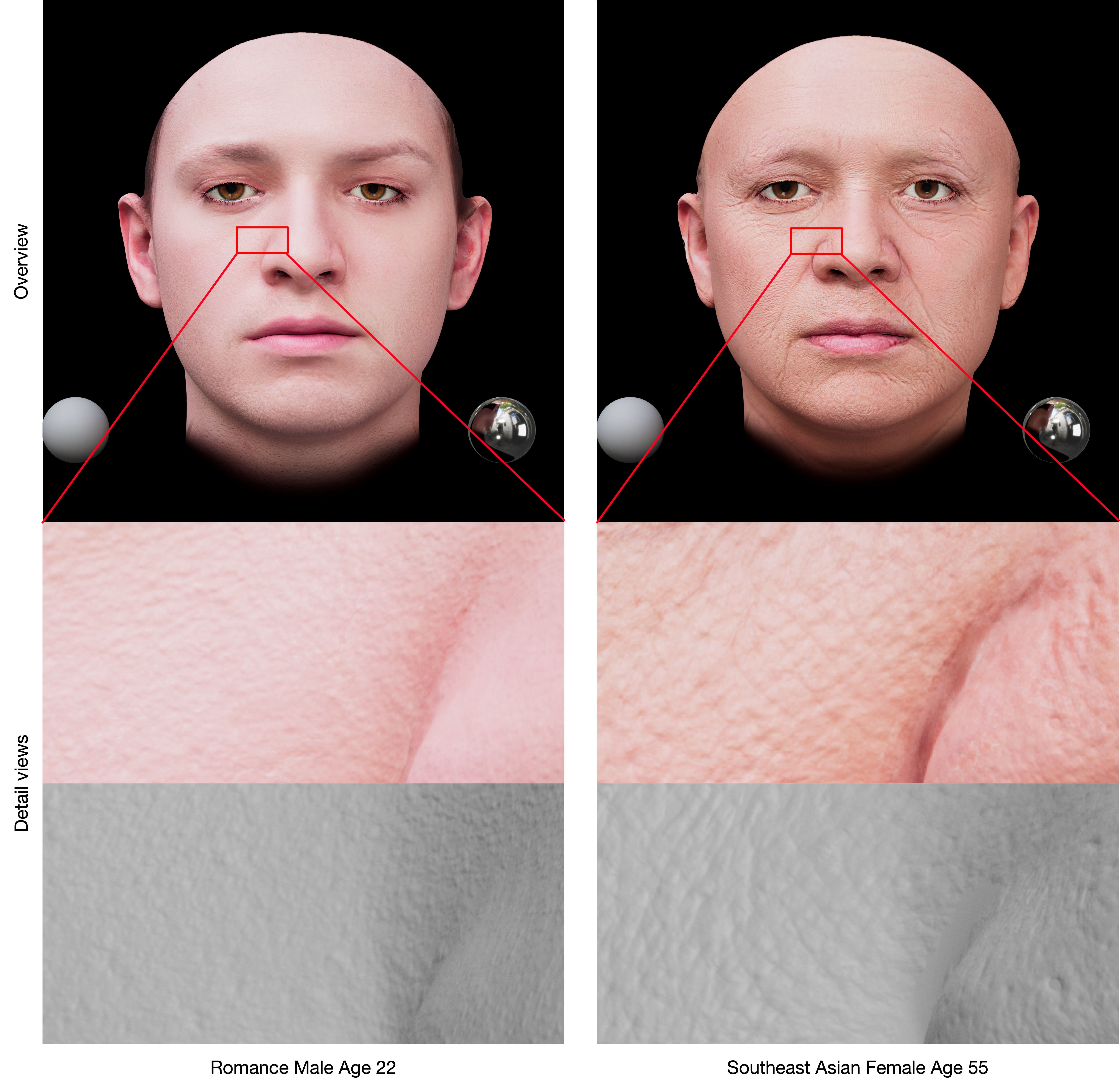}
\caption{Zoom-in views of generated face models. Two different subjects are shown. Top: full facial renderings with highlighted regions. Bottom: close-ups showing details of highlighted regions in the albedo map and geometry.}
\Description{}
\label{fig:res_detail}
\end{figure*}

\begin{figure}
\centering
\includegraphics[width=0.8\linewidth]{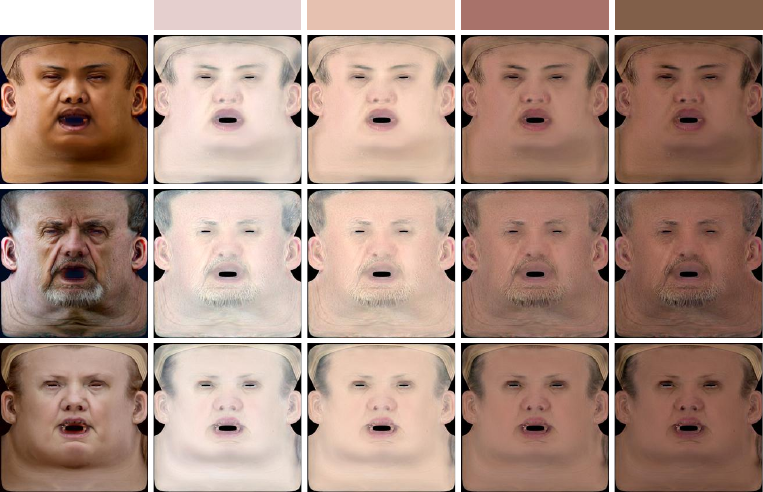}
\caption{Skin color control in texture normalization.}
\Description{}
\label{fig:norm_color}
\end{figure}

\begin{figure}
\centering
\includegraphics[width=0.9\linewidth]{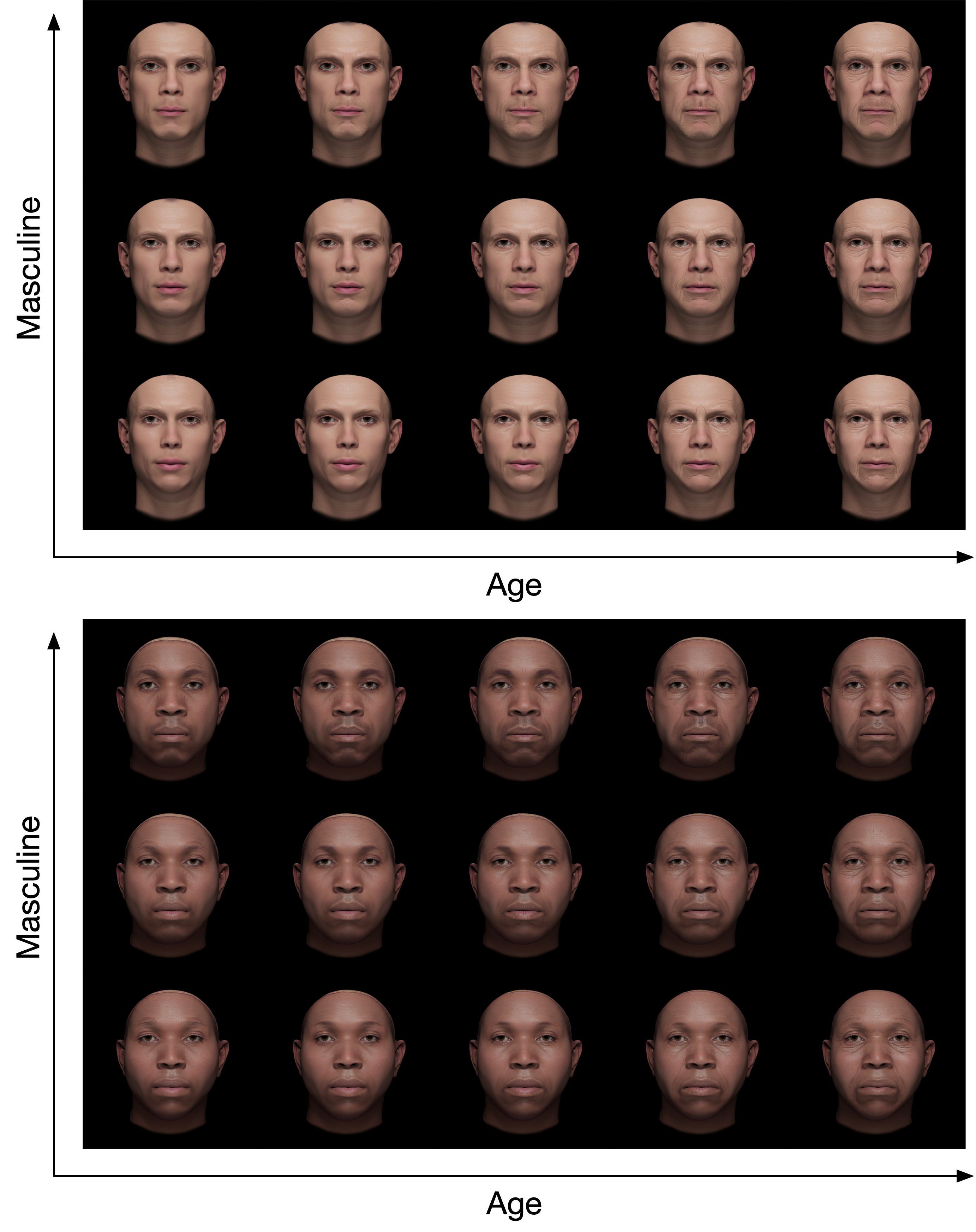}
\caption{Independent control of age and gender. Each row varies gender from feminine to masculine, and each column increases age progressively.}
\Description{}
\label{fig:ctrl_ag}
\end{figure}

\begin{figure}
\centering
\includegraphics[width=0.35\linewidth]{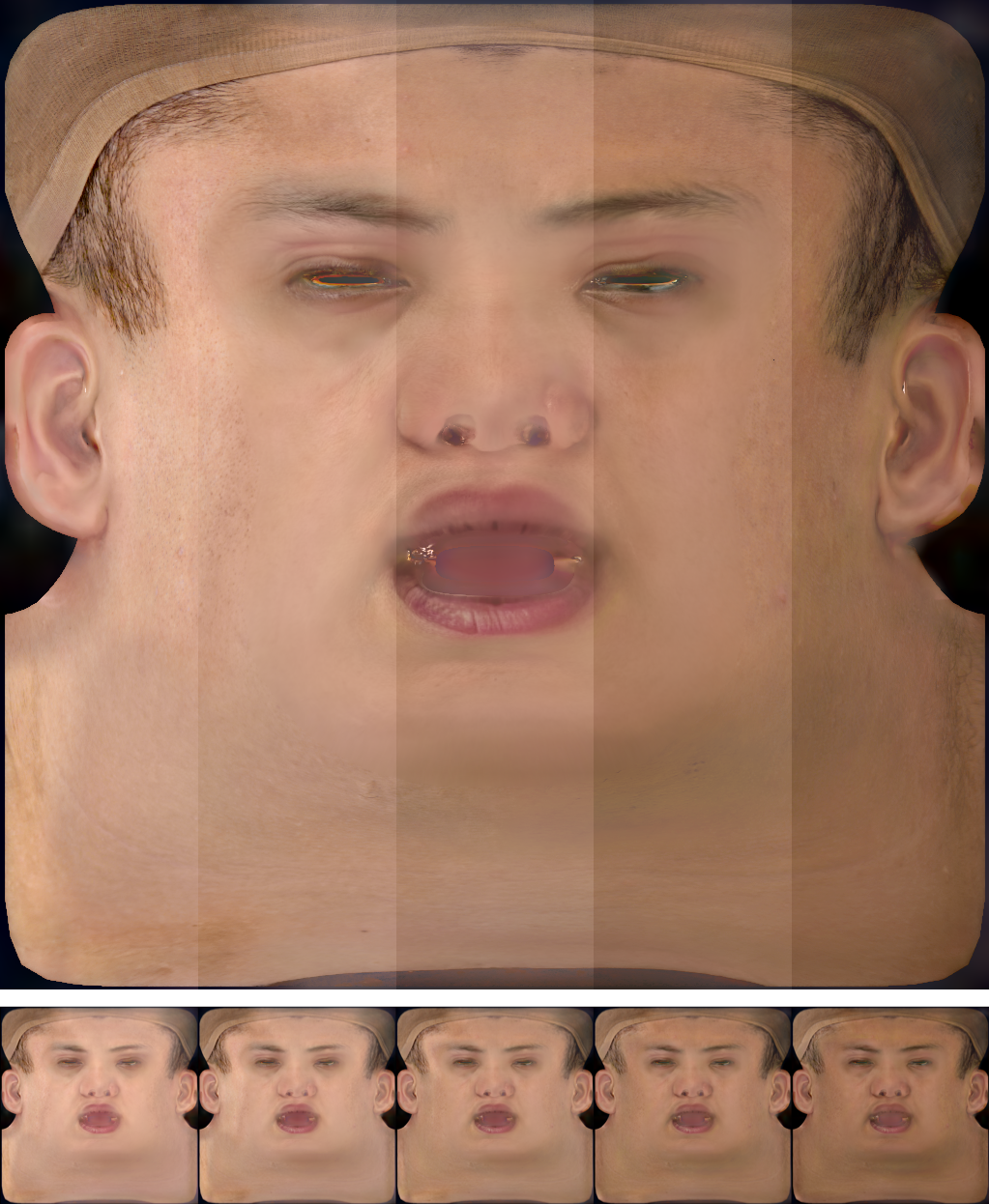}
\hspace{0.05\linewidth}
\includegraphics[width=0.35\linewidth]{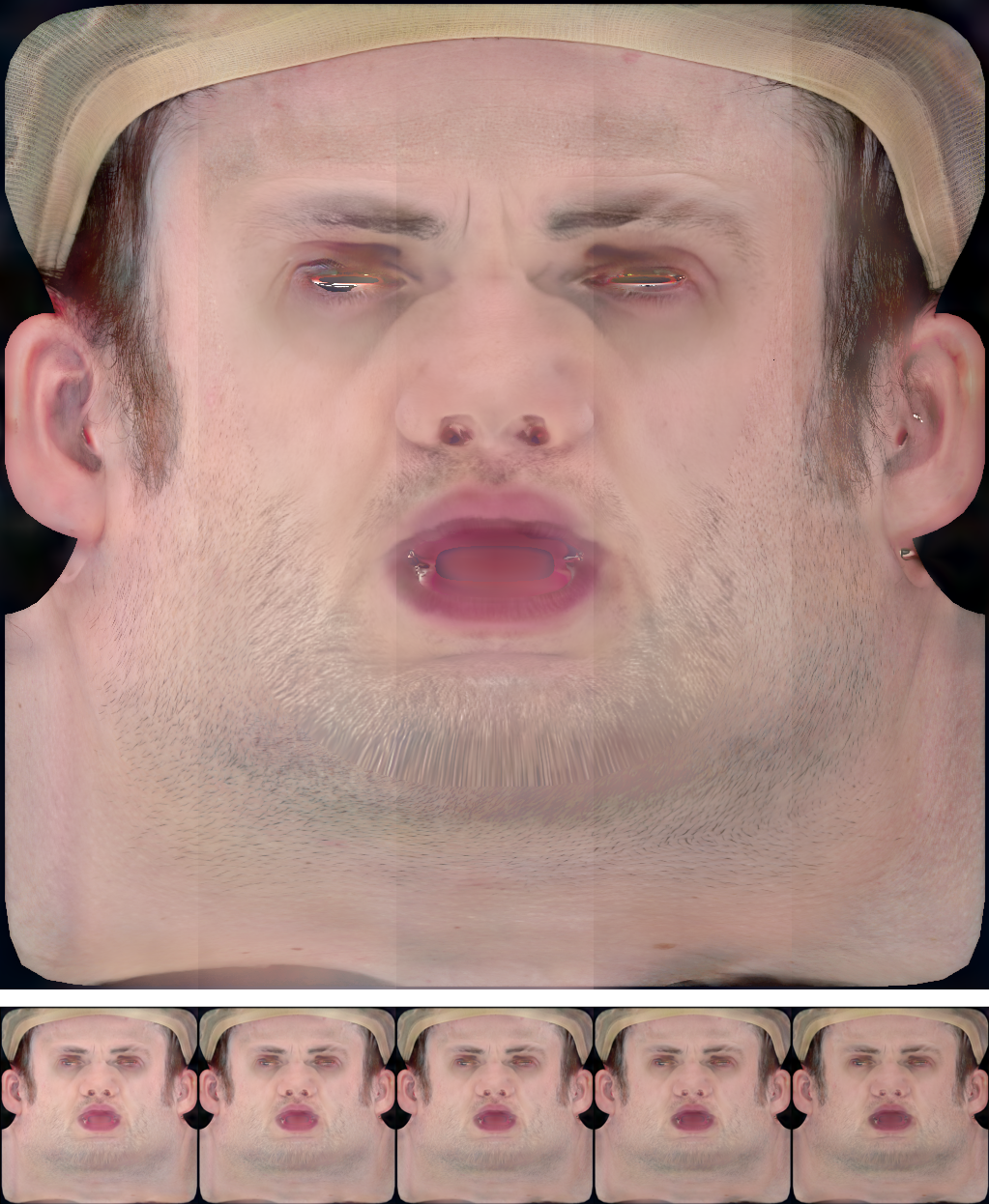}
\vspace{0.5em}
\makebox[0.35\linewidth]{\small South Asian Female Age 28}
\hspace{0.05\linewidth}
\makebox[0.35\linewidth]{\small Germanic Unisex Age 40}
\caption{Skin color control in image generation.}
\Description{}
\label{fig:ctrl_c}
\end{figure}

\begin{figure}
\centering
\includegraphics[width=\linewidth]{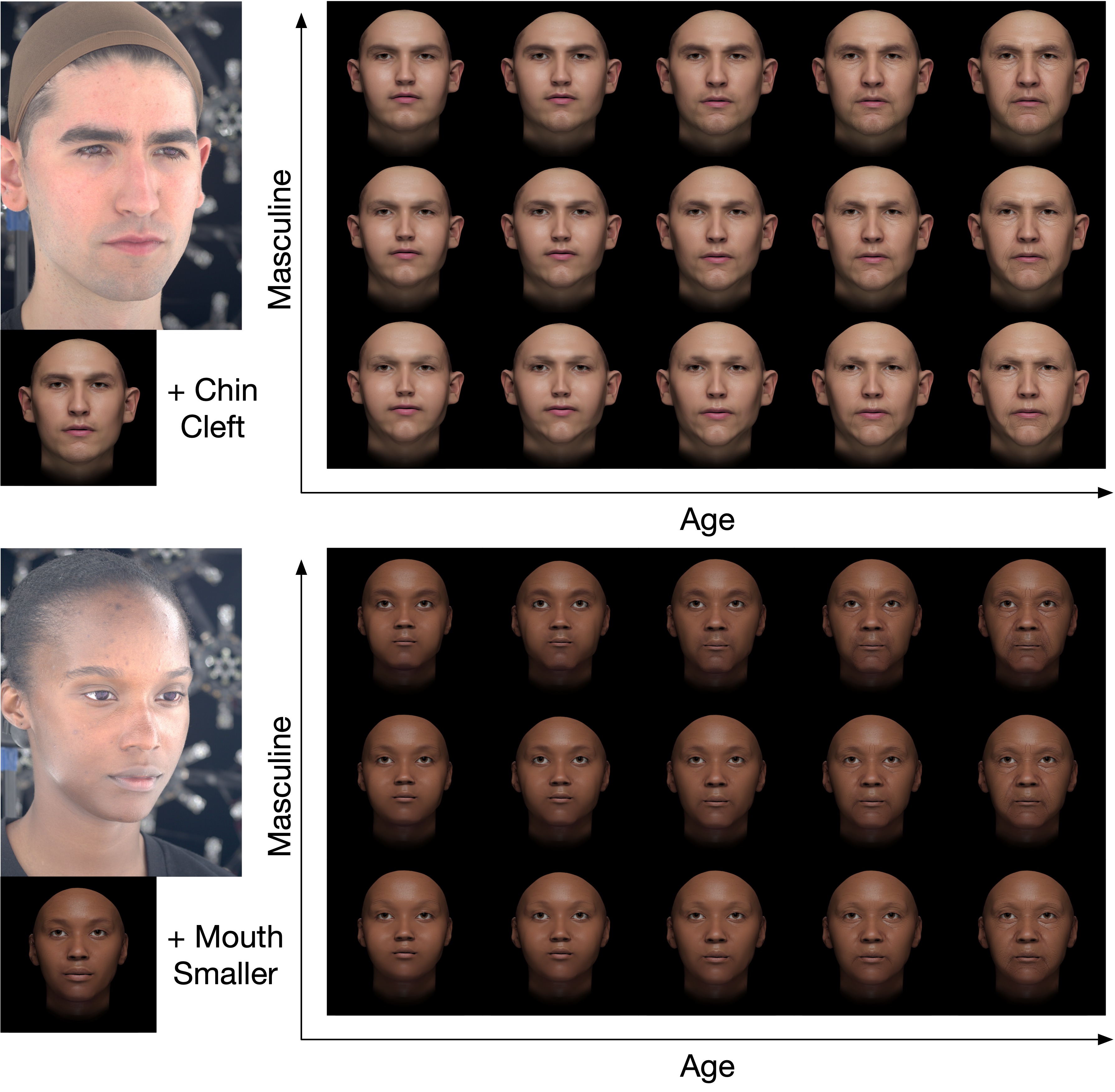}
\caption{Examples of GAN inversion and editing of facial features. Each subject is inverted from a real photo, then modified on the geometry—adding a chin cleft (top) or reducing mouth size (bottom)—followed by age and gender editing in the latent space of the generator.}
\Description{}
\label{fig:gan_inversion}
\end{figure}

\begin{figure}
\centering
\includegraphics[width=\linewidth]{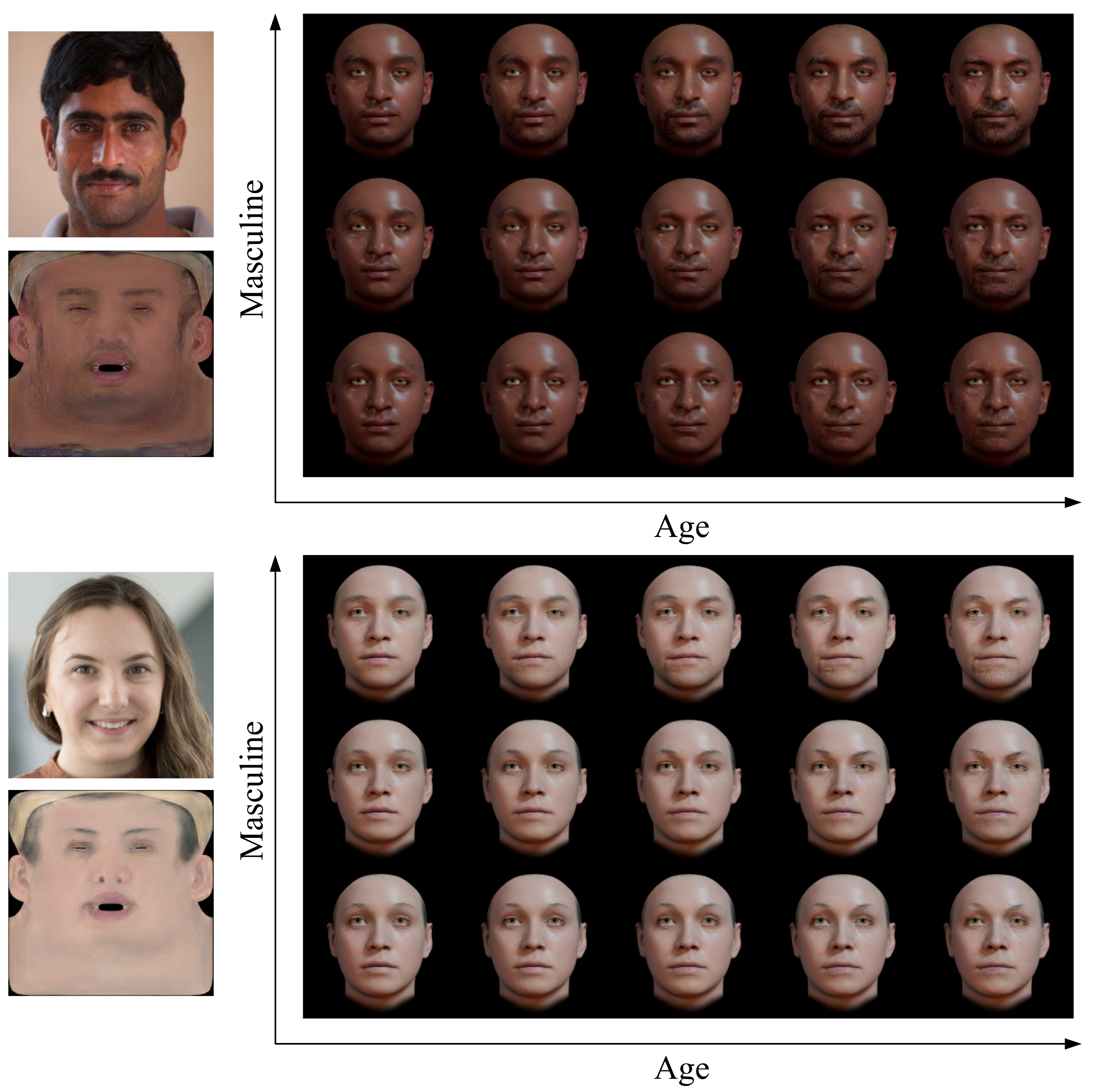}
\caption{Examples of GAN inversion and editing on in-the-wild images, from the FFHQ dataset. The texture shown is the result of inversion.}
\Description{}
\label{fig:gan_inversion_ffhq}
\end{figure}

\subsection{Results}
\subsubsection{Qualitative Results}
\paragraph{Face Model Generation}
\Cref{fig:res_assets} shows full assets of 16 identities along with rendered images, created from 8 distinct semantic labels. For each row, two randomly generated results are shown (column 2-5, column 6-9) that follow the same input semantic attributes under different identities. For each subject, the geometry, diffuse albedo map, specular map, and displacement map are created from a set of attributes (1st column) and rendered under three different lighting conditions. These results illustrate the diversity, quality, and effective semantic control of our model.

\paragraph{Texture normalization.}
The result of texture normalization is shown as part of \Cref{fig:normarlizationresults}, where columns 3 and 4 show training data before and after normalization. Skin color is given in column 1. It transforms the UV-space texture with baked-in illumination into albedo textures while preserving the identity of the input texture. \Cref{fig:norm_color} shows the effect of skin color control in texture normalization.

\paragraph{Attribute Control.}
\Cref{fig:ctrl_ag} shows independent control of age and gender in both texture and geometry. In each block, gender varies within each column and age varies within each row, while all other attributes remain fixed. Along the gender axis, we can see that a masculine face generally displays sharper, more pronounced features, including thicker eyebrows. Along the age axis, the most notable difference is the prominence of wrinkles.

\Cref{fig:ctrl_c} shows independent control of skin color in the generator. When skin color is not specified, it is sampled randomly from the skin color distribution of the chosen ethnicity, which is a normal distribution fit to the skin color distribution in the unnormalized training data. The figure presents five skin color samples of each subject, from bright to dark in the range of their ethnicity, while keeping identity and facial features consistent.

\paragraph{Inversion and Editing.}
Figures \ref{fig:gan_inversion} and \ref{fig:gan_inversion_ffhq} shows image inversion and editing. Images are first projected into the latent space of the generator by finding the unlabeled code and labeled attributes that generates the closest image using optimization. Edited images are then generated by modifying the labeled attributes while keeping the unlabeled code unchanged.

Additionally, since all generated facial geometries share the same topology from our 3D asset dataset $\mathcal{D}_{main}$, generic facial geometric attributes editing such as add aging features \Cref{fig:ctrl_ag} can be done with some predefined geometry offsets, and blendshapes can be applied to animate the facial assets. For animation examples, please refer to the video.

\subsubsection{Comparative Evaluation}

\paragraph{Texture normalization.}
For a quantitative evaluation, we can take a UV-space image rendered using scanned data, normalize it, and compare the result to the ground truth albedo by calculating the PSNR. We compare our method with FFHQ-UV~\cite{bai2023ffhq}. However, in FFHQ-UV, the facial geometry fitting and texture normalization are integrated, and it cannot be applied to UV textures directly. So, we render our scanned data and apply FFHQ-UV. The result is in \Cref{tab:psnr}. 

In addition, on datasets without known ground truth we calculate the Brightness Symmetry Error (BS Error) introduced in FFHQ-UV, as shown in \Cref{tab:bs_error}. The metric measures the dissimilarity of brightness between the left and right halves of the face, based on the assumption that without lighting the brightness of the two halves should be identical. For the Facescape data, we processed it using a method similar to~\cite{li2020learning} to obtain texture data with the same topology. For FFHQ, we employed a 3DMM fitting method to obtain partial textures from the images and then applied the blending described in \Cref{sec:text_comp} to complete the textures around the central facial area. For a qualitative evaluation, \Cref{fig:ffhquv} shows the results of texture normalization compared to FFHQ-UV~\cite{bai2023ffhq}.
Our normalization results are closer to the target albedo domain, with pure skin color and no lighting baked-in (e.g., highlights). However, the normalized UV generated by FFHQ-UV still retains some of the original illumination, which cannot be used for high-end PBR-base shader as albedo.  

\begin{table}[h]
\centering
\caption{Quantitative evaluation of texture normalization using PSNR. }
\label{tab:psnr}

\begin{tabular}{ccc}
\toprule
Method & FFHQ-UV & Ours \\
\midrule
PSNR $\uparrow$ & 17.64 & 24.67 \\
\bottomrule
\end{tabular}
\end{table}

\begin{table}[htbp]
\centering
\caption{Quantitative evaluation on the illumination of the proposed UV-texture dataset in terms of BS Error, where * denote the dataset which is captured under controlled conditions, and ** indicate evaluation on face models with varying topology. Ours refers to FFHQ processed by proposed normalization method to match Light Stage data. }
\label{tab:bs_error}

\begin{tabular}{@{}lccccc@{}}
\toprule
Method & \makecell{Facescape*} & \makecell{Light\\Stage*} & FFHQ & \makecell{FFHQ-\\UV**} & Ours \\
\midrule
BS Error $\downarrow$ & 15.2595 & 4.8279 & 28.0723 & 7.293 & 5.8109 \\
\bottomrule
\end{tabular}
\end{table}

\begin{figure*}
\centering
\includegraphics[width=0.95\linewidth]{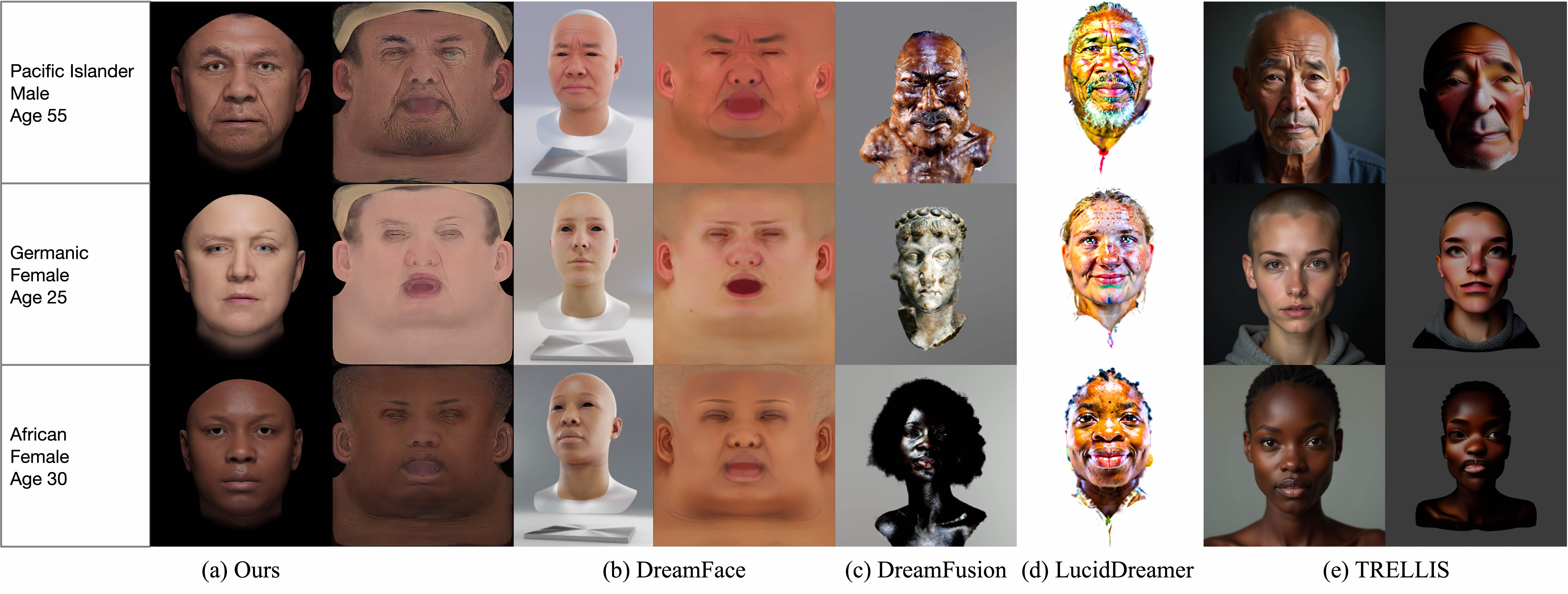}
\caption{Comparison of semantic face asset generation results of (a) the proposed method against (b) DreamFace \cite{zhang2023dreamface}, (c) DreamFusion \cite{poole2022dreamfusion}, (d) LucidDreamer \cite{EnVision2023luciddreamer}, and (e) TRELLIS \cite{xiang2024structured}. }
\Description{}
\label{fig:res_dreamface}
\end{figure*}

\paragraph{Facial Attribute Control.}
\Cref{fig:res_dreamface} compares our system to DreamFace ~\cite{zhang2023dreamface}, a state-of-the-art face asset generation approach. Our system generates real albedo, which reflects true skin color, whereas DreamFace only produces texture under evenly distributed illumination. Also, our generated face model exhibits greater diversity and more closely aligns with the user description as the training dataset of DreamFace mainly consists of Asian people. Compared to general text-to-3D methods like DreamFusion~\cite{poole2022dreamfusion}, LucidDreamer~\cite{EnVision2023luciddreamer} and TRELLIS~\cite{xiang2024structured}, our method produces avatars with impressive detail and production-level textures. DreamFusion creates clear structures, but misses finer skin details. LucidDreamer and TRELLIS output vivid, detailed and realistic faces, but with strong artifacts. 

\begin{figure}
\centering
\includegraphics[width=0.4\linewidth]{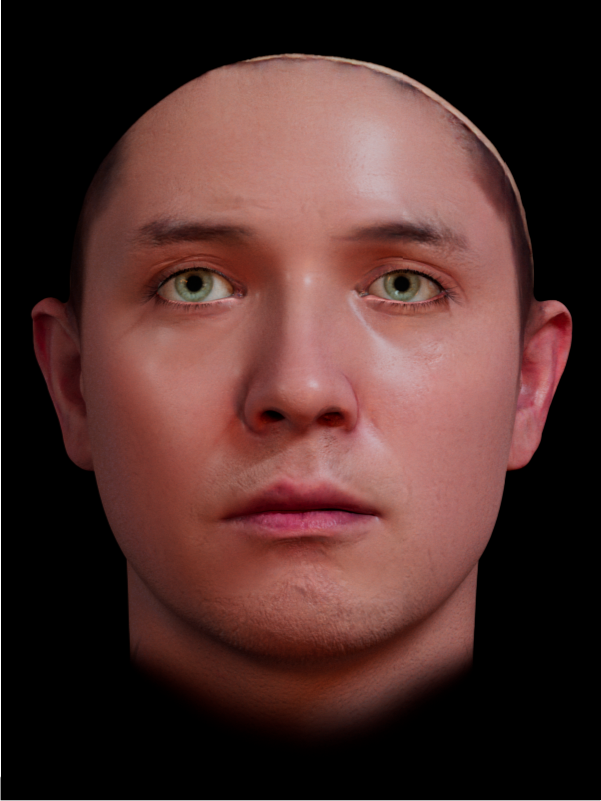}
\hspace{0.05\linewidth}
\includegraphics[width=0.4\linewidth]{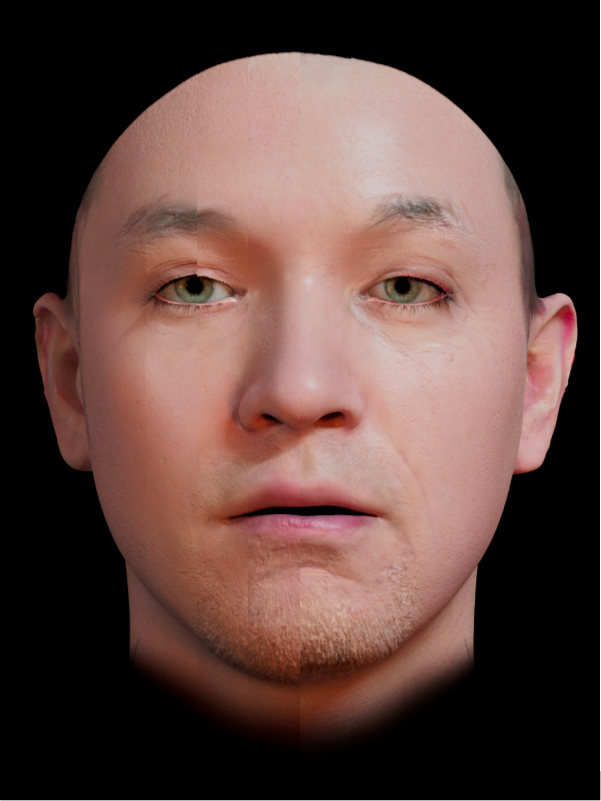}

\vspace{0.2em}
\makebox[0.4\linewidth]{\small Ours}
\hspace{0.05\linewidth}
\makebox[0.4\linewidth]{\small Diffusion}

\caption{Age editing comparison between our method and text-prompt-based editing using Stable Diffusion~\cite{rombach2022high}. For each method, 20 ages ranging from young to old are uniformly sampled. Each generated face contributes a vertical slice.}
\Description{}
\label{fig:comp_edit}
\end{figure}

To compare the effectiveness of attribute control between our trained generator and a generic diffusion model~\cite{rombach2022high}, we perform age editing on the same subject using both methods, as shown in~\Cref{fig:comp_edit}. The portraits generated by the diffusion model, which is the same as the one used in portraits generation, are processed through our data preparation pipeline to obtain final renderings. While diffusion models can achieve age modification through prompt manipulation, our generator produces significantly smoother transitions across age levels. In contrast, diffusion-based results have abrupt feature changes between adjacent slices in the figure. \textit{A more detailed comparison of transition smoothness can be found in the supplementary video}.

\begin{figure}
\centering
\includegraphics[width=1.0\linewidth]{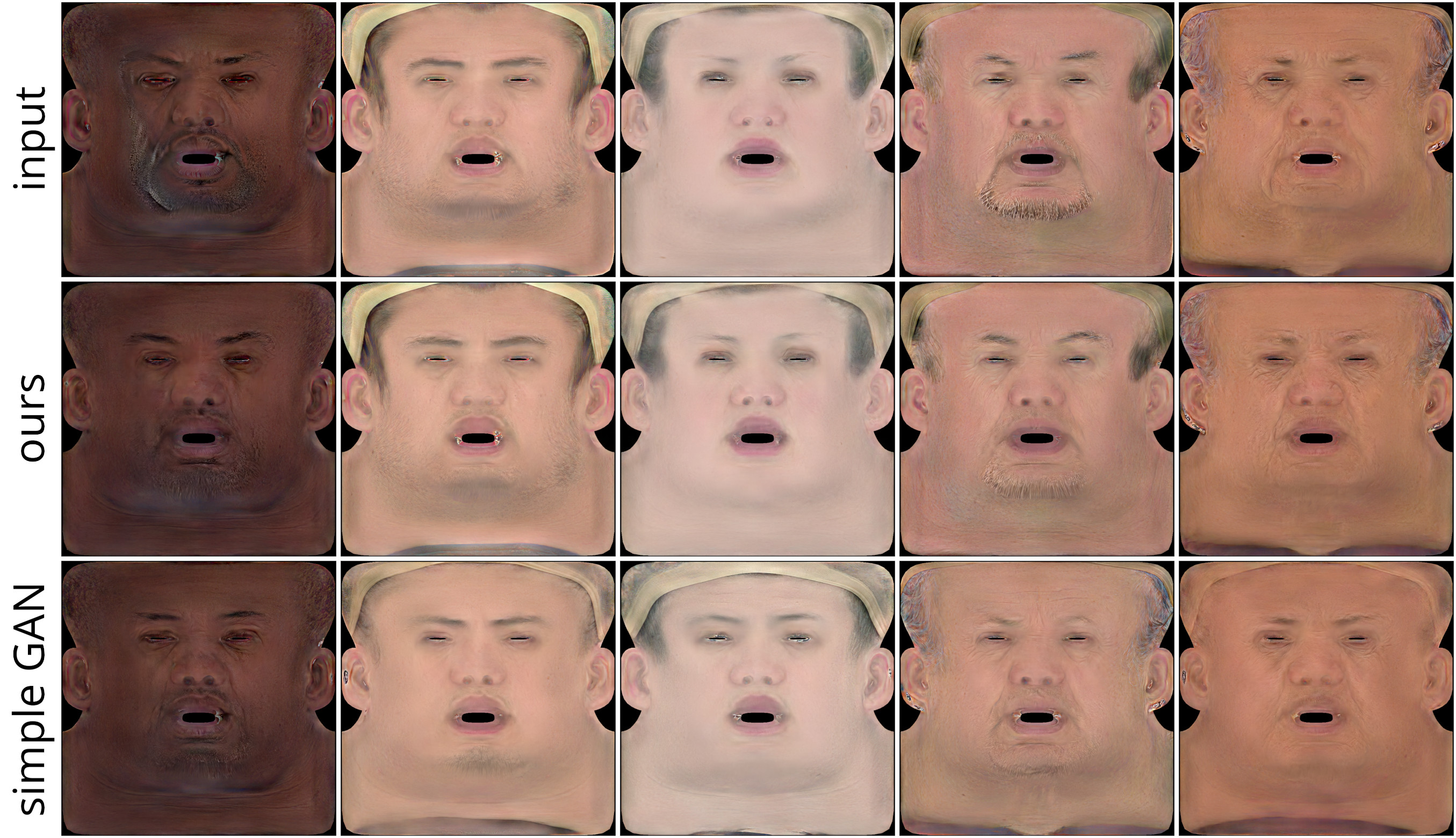}
\caption{Comparison of image inversion using our generator versus a simple conditional GAN.}
\Description{}
\label{fig:gan_inversion_comparison}
\end{figure}

For a quantitative comparison of attribute control of our generator to the other methods, we use CLIP score. We first build a text description of the generated image data for our semantic attributes. We utilize the same ethnicity, gender, and age groups as the attributes we defined for generation to construct descriptors. Following the strategy of~\cite{zhang2023dreamface}, we form text input with the phrase ``the realistic face of [DESCRIPTOR]''. After acquiring 5 sets of PBR assets for each descriptor, we render 2500 test images. To calculate the score of Describe3D~\cite{wu2023high}, we synthesize facial images by generating the descriptors with the rules mentioned in their work and used the same phrase as input into the CLIP model. Subsequently, we compute the average CLIP score from the ViT-B/16 and ViT-L/14 models. The result, shown in \Cref{tab:comparison}, highlights the effectiveness of our method in achieving semantic coherence between the generated images and their corresponding text descriptions.

\begin{table}[t]
\centering
\caption{Comparison of CLIP score to baseline Methods.}
\label{tab:comparison}
\begin{tabular}{lcc}
\hline
\textbf{Method} & \textbf{CLIP Score $\uparrow$} \\
\hline
DreamFace~\cite{zhang2023dreamface} & 0.291 ± 0.020 \\
Describe3D~\cite{wu2023high} & 0.284 ± 0.054 \\
UltrAvatar~\cite{zhou2024ultravatar} & 0.301 ± 0.023 \\
\textbf{Our Method} & \textbf{0.316 ± 0.042} \\
\hline
\end{tabular}
\end{table}

\begin{figure}
    \centering
    \begin{subfigure}[b]{0.48\textwidth}
        \centering
        \includegraphics[width=\textwidth]{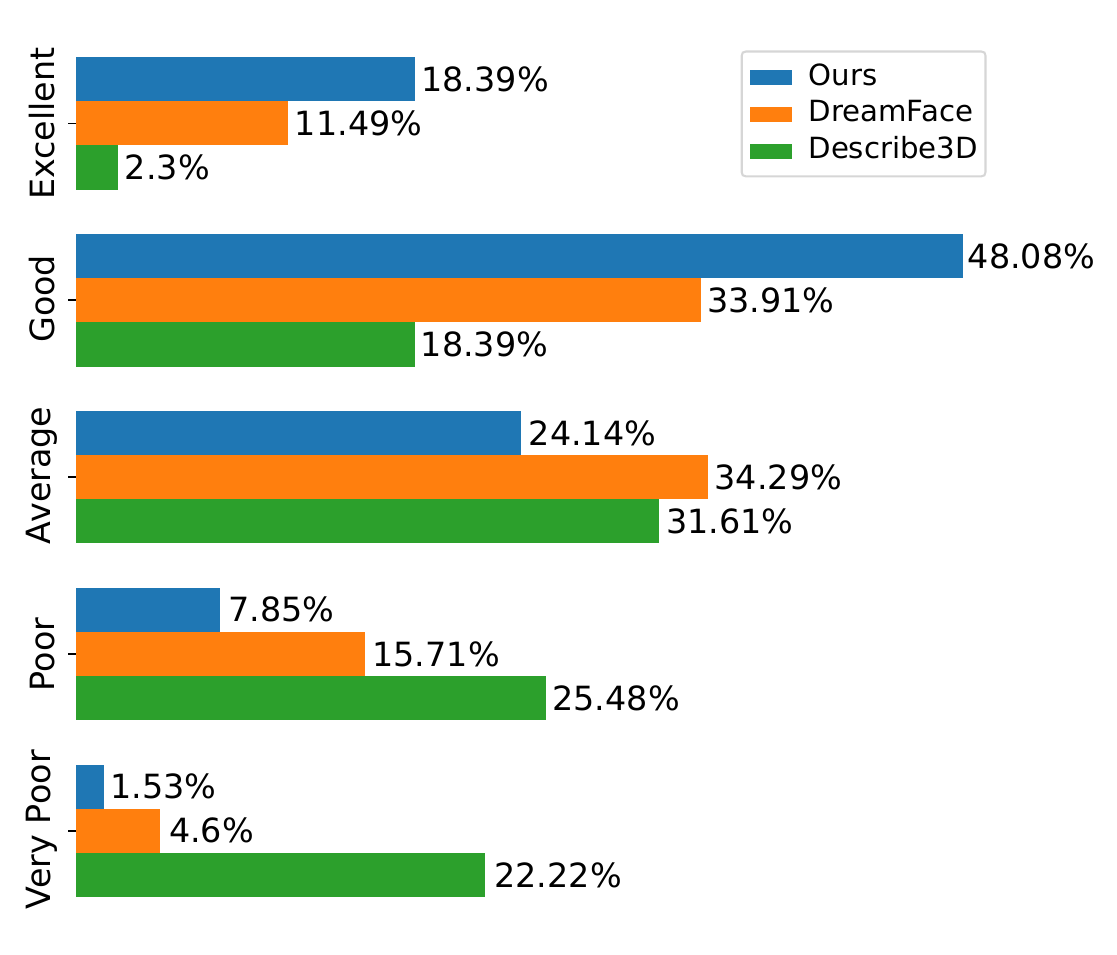}
        \caption{Description Consistency}
        \Description{}
    \end{subfigure}
    \hfill
    \begin{subfigure}[b]{0.48\textwidth}
        \centering
        \includegraphics[width=\textwidth]{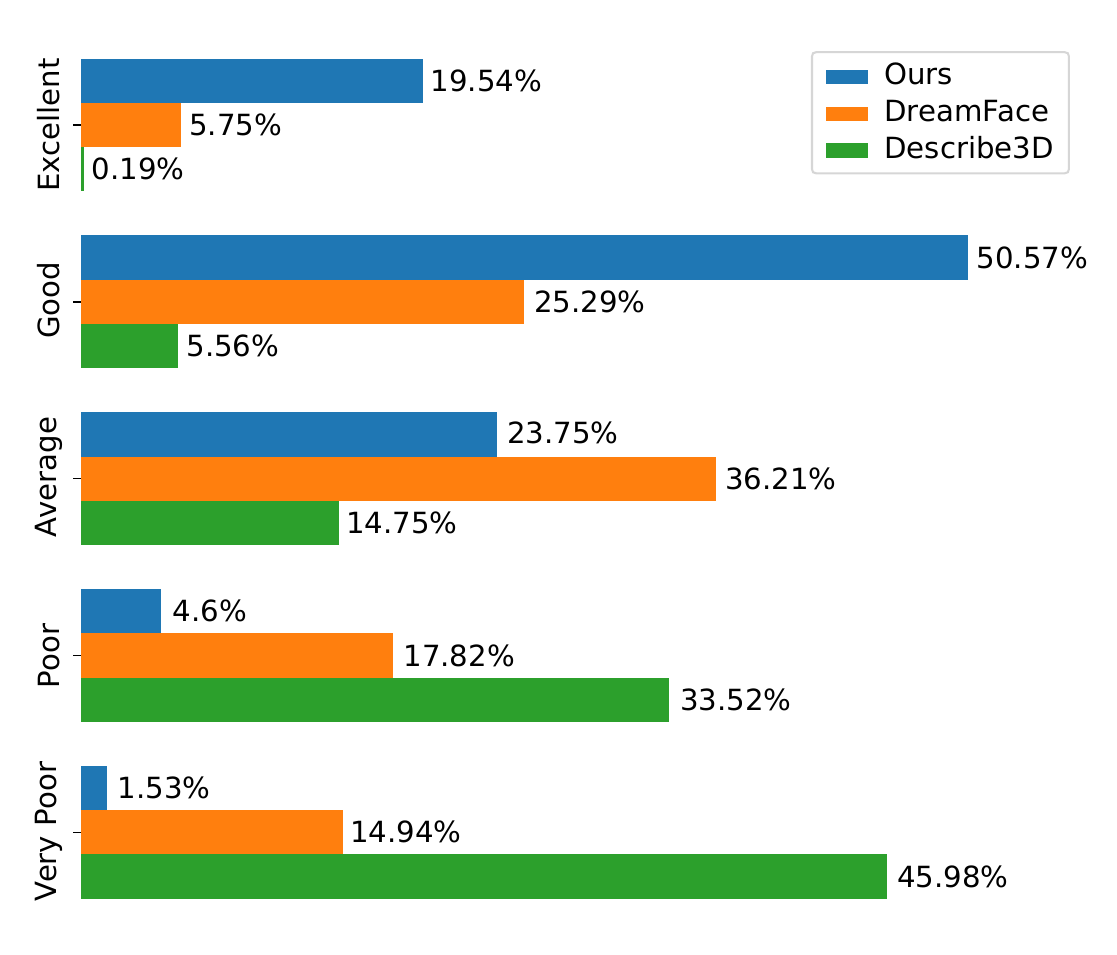}
        \caption{Photorealism}
    \end{subfigure}
    \caption{Comparison of different methods across various categories}
    \Description{}
    \label{fig:userstudy}
\end{figure}

\paragraph{User Study} We conduct a user study to compare the outputs of avatar generation methods, including our proposed method, DreamFace~\cite{zhang2023dreamface} and Describe3D~\cite{wu2023high}. 87 participants rate images synthesized using the corresponding methods in two aspects: description consistency and photorealism. The outcomes in \Cref{fig:userstudy} show that our proposed method consistently outperforms the baseline methods in terms of description consistency and photorealism.

\subsubsection{Ablation Study}
\label{sec:ablation}

\begin{figure}
\centering
\includegraphics[width=\linewidth]{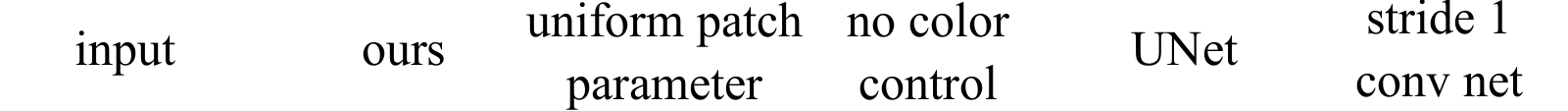}
\includegraphics[width=\linewidth]{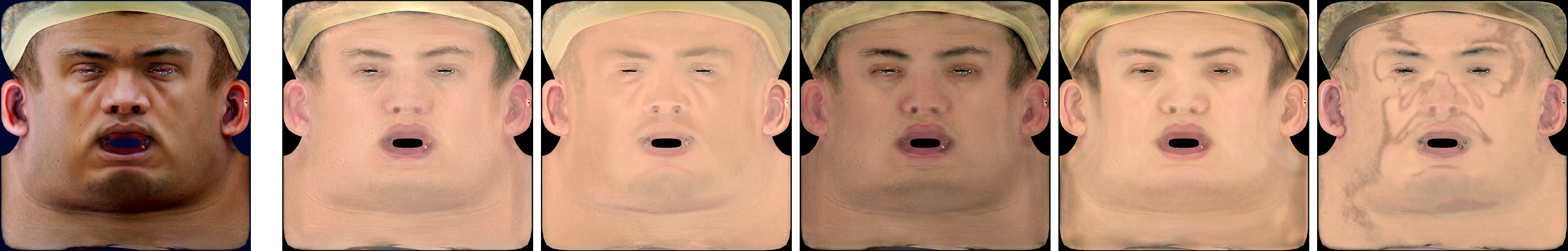}
\includegraphics[width=\linewidth]{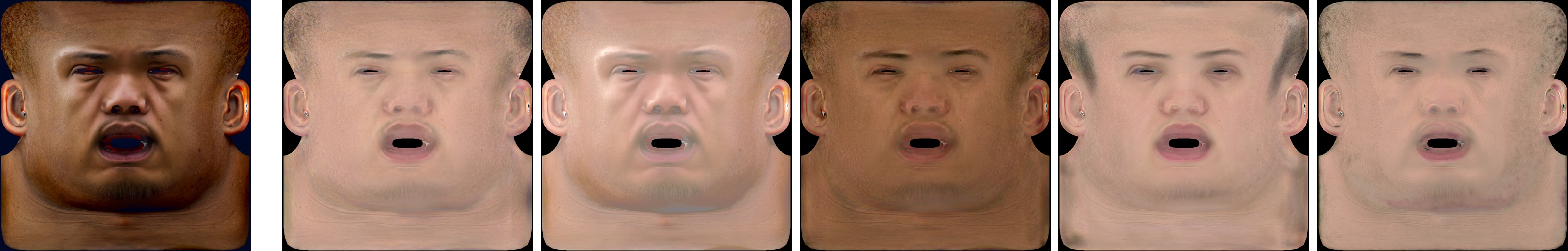}
\includegraphics[width=\linewidth]{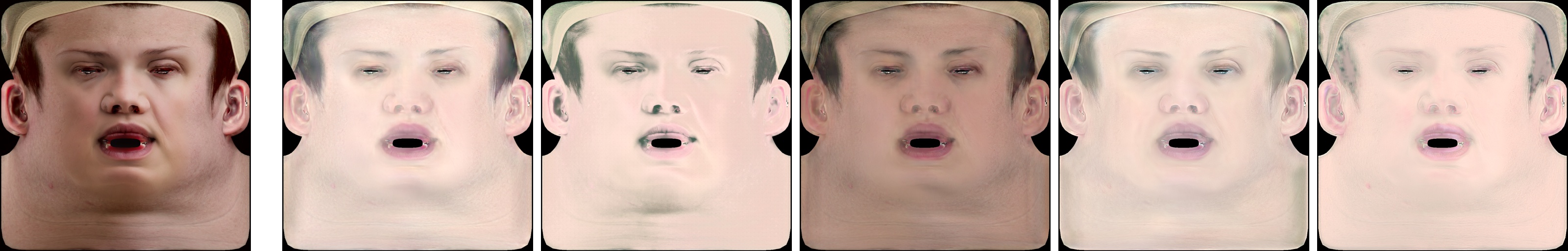}
\includegraphics[width=\linewidth]{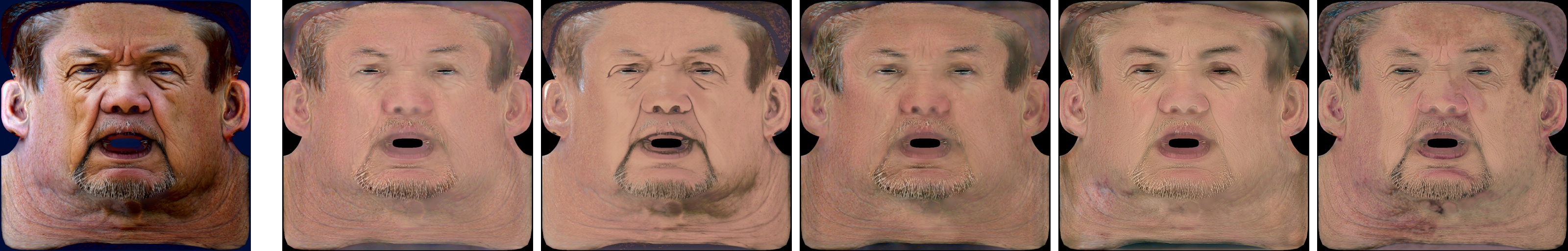}
\caption{Ablation study for the Texture Normalization Network. Each row shows prediction of UV-space albedo maps produced by five different network architecture designs for the normalization network (see Section~\ref{sec:ablation}) for details.}
\Description{}
\label{fig:norm_abl}
\end{figure}

\paragraph{Design of the Texture Normalization Network.}
To verify our design for the normalization network, we compare it with the following variations: 1) \textit{``uniform patch parameter''}: removing the spatial variability of patch parameter by pooling the output of the patch parameter network and broadcasting it to every location, 2) \textit{``no color control''}: removing external color control and instead using the color calculated from the flat regions of the input image, 3) \textit{``Unet''}: replacing the normalizer with a UNet, 4) \textit{``stride 1 conv net''}: replacing the normalizer with a simple convolution network containing only stride-1 convolutions. The results are shown in Figure \ref{fig:norm_abl}. The network with uniform patch parameters only applies a global color mapping. Due to the invariability of patch parameters it cannot handle strong local variations in lighting, as shown in row 2. Without color control, the quality is comparable, but the model loses its ability to incorporate color information if such information is available. In these examples, the race is known, and the color input is drawn randomly from the color distribution of the race, and the network without color control fails to produce race-accurate output skin color. The UNet enables long-range exchange of information and could modify facial features if that results in a better match between the output normalized texture and the high-quality scan. The scanned data contains a large portion of subjects with heavy eyebrows, and it can be observed that in row 4, the eyebrow appears darker and more expansive in the UNet output than the ground truth. In the stride-1 convnet, as each output pixel depends only on a small patch in the input, there is no coordination across the image, which can result in major failure cases as in row 1. Overall, the ablation study validated that our normalization design performs the best compared to the alternatives.

\begin{figure}
\centering
\begin{tabular}{cccc}
    \rule[-8pt]{0pt}{10pt} & race & gender & age \\
    \rotatebox{90}{\hspace{17pt}dataset} &
    \includegraphics[width=0.25\linewidth]{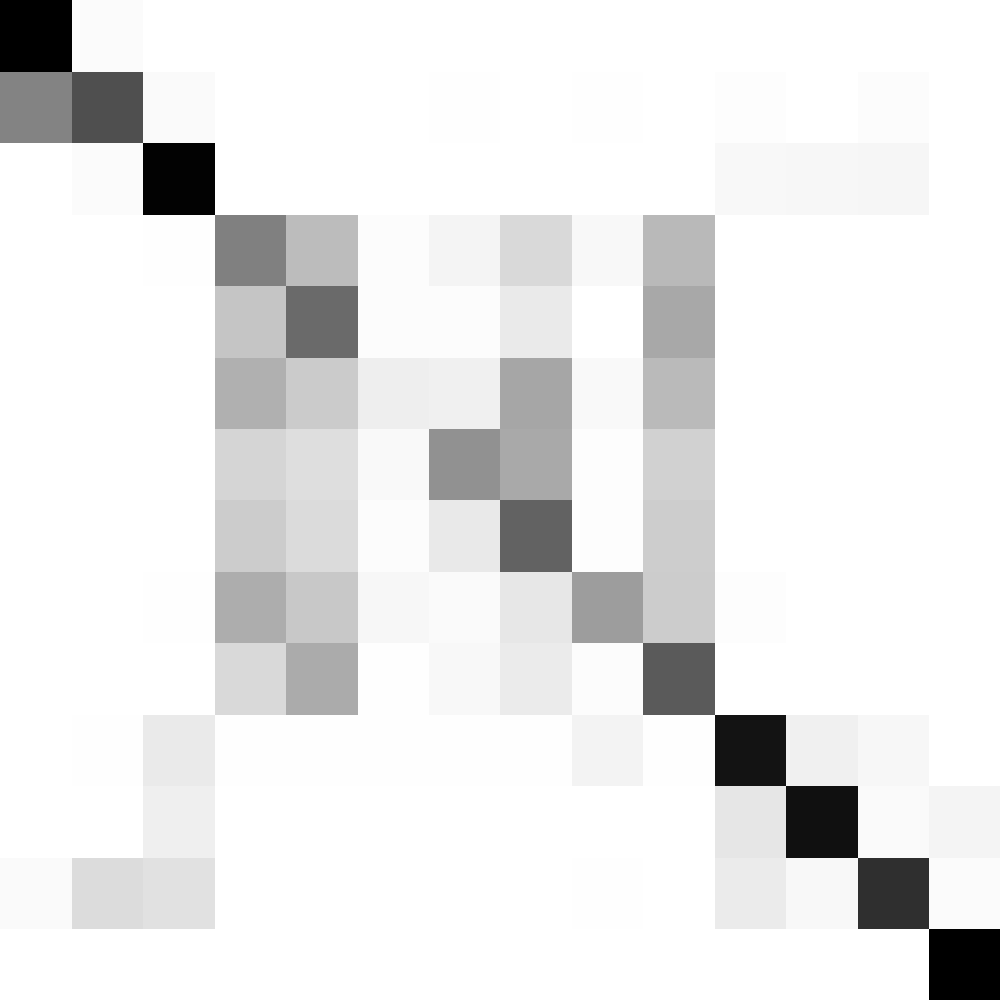} &
    \includegraphics[width=0.25\linewidth]{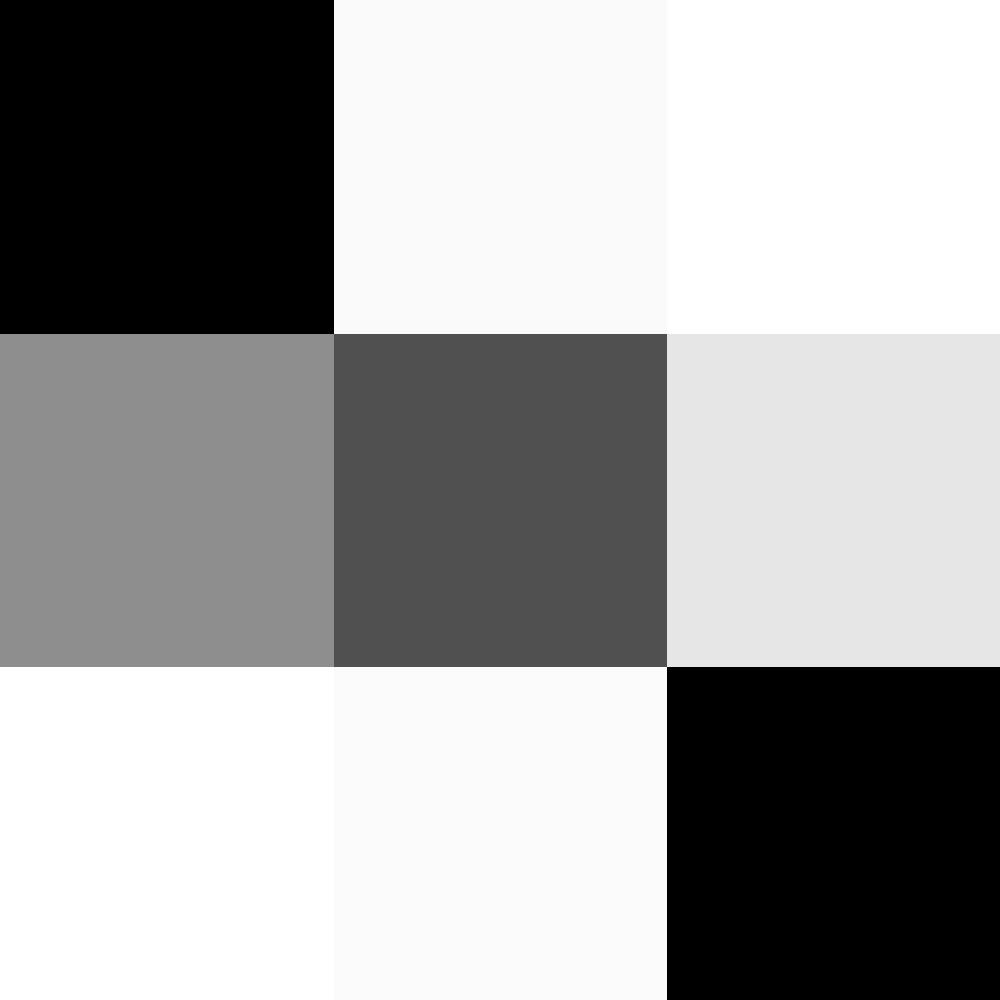} &
    \includegraphics[width=0.25\linewidth]{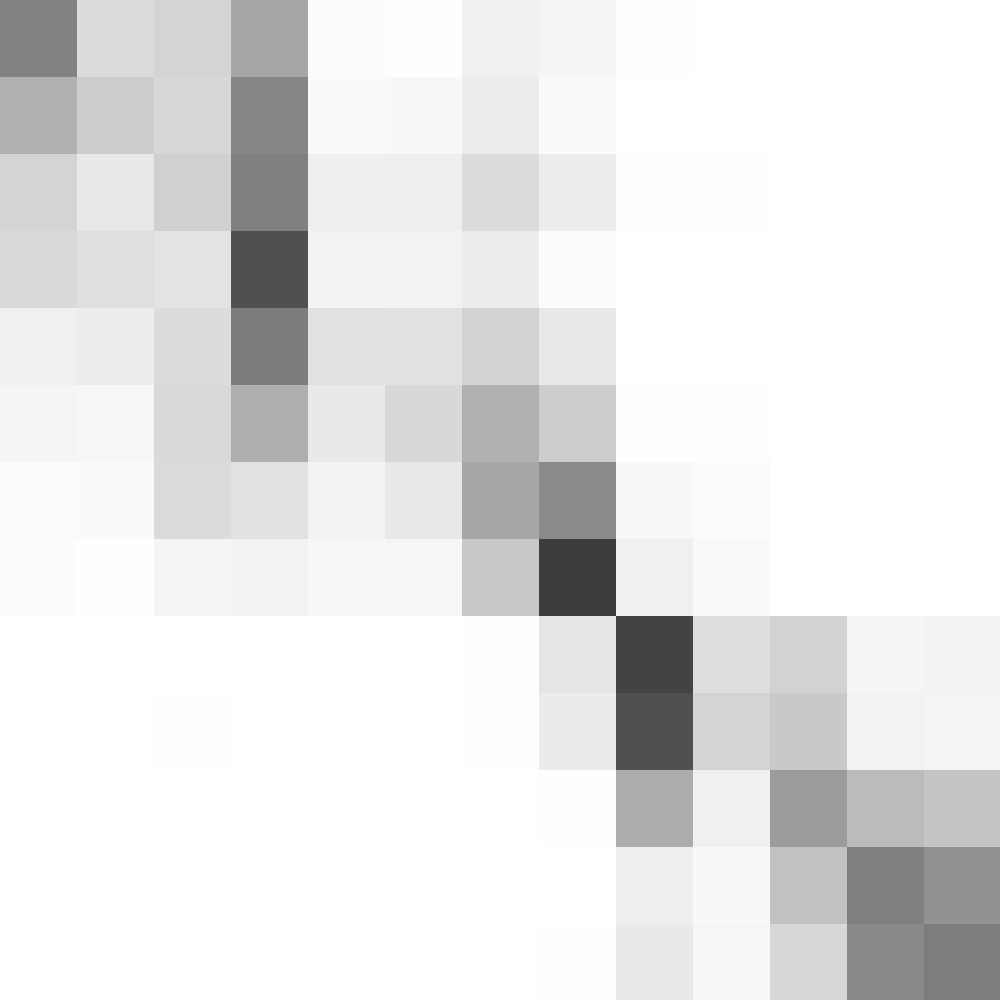} \\
    \rotatebox{90}{\hspace{12pt}generated} &
    \includegraphics[width=0.25\linewidth]{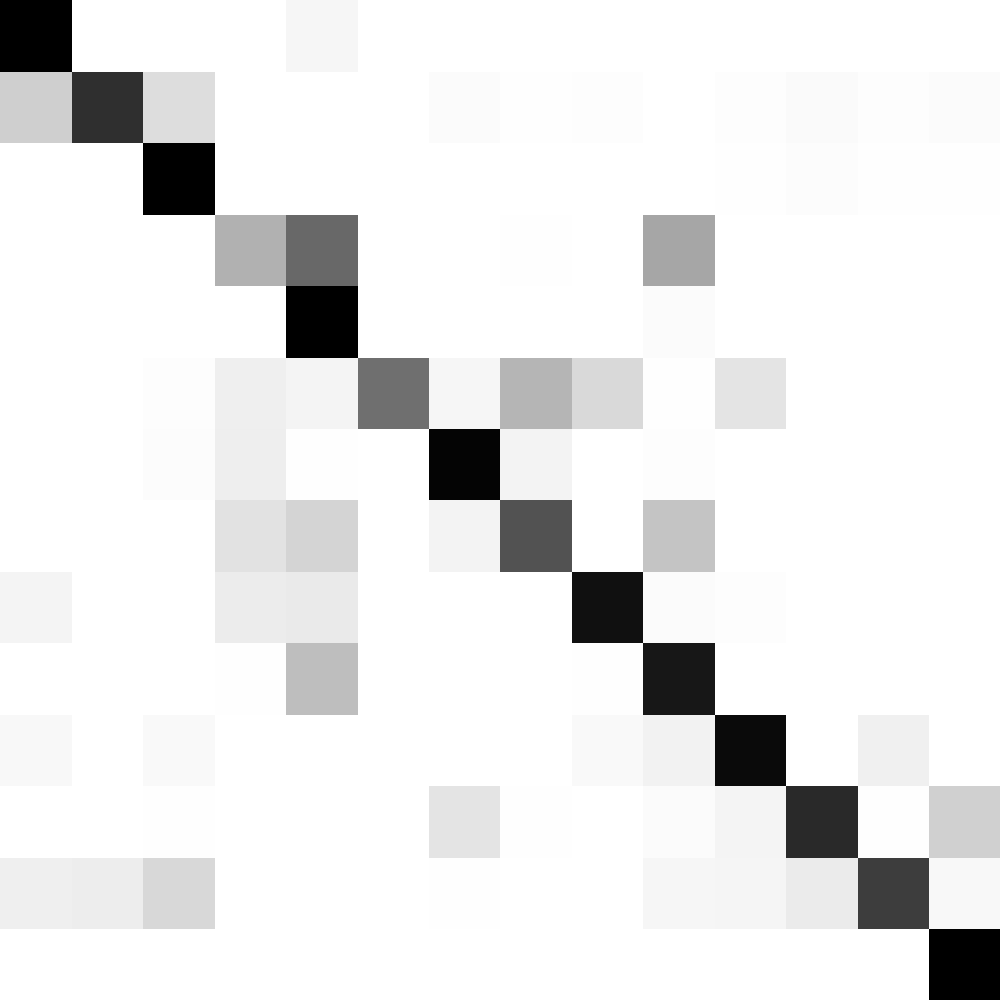} &
    \includegraphics[width=0.25\linewidth]{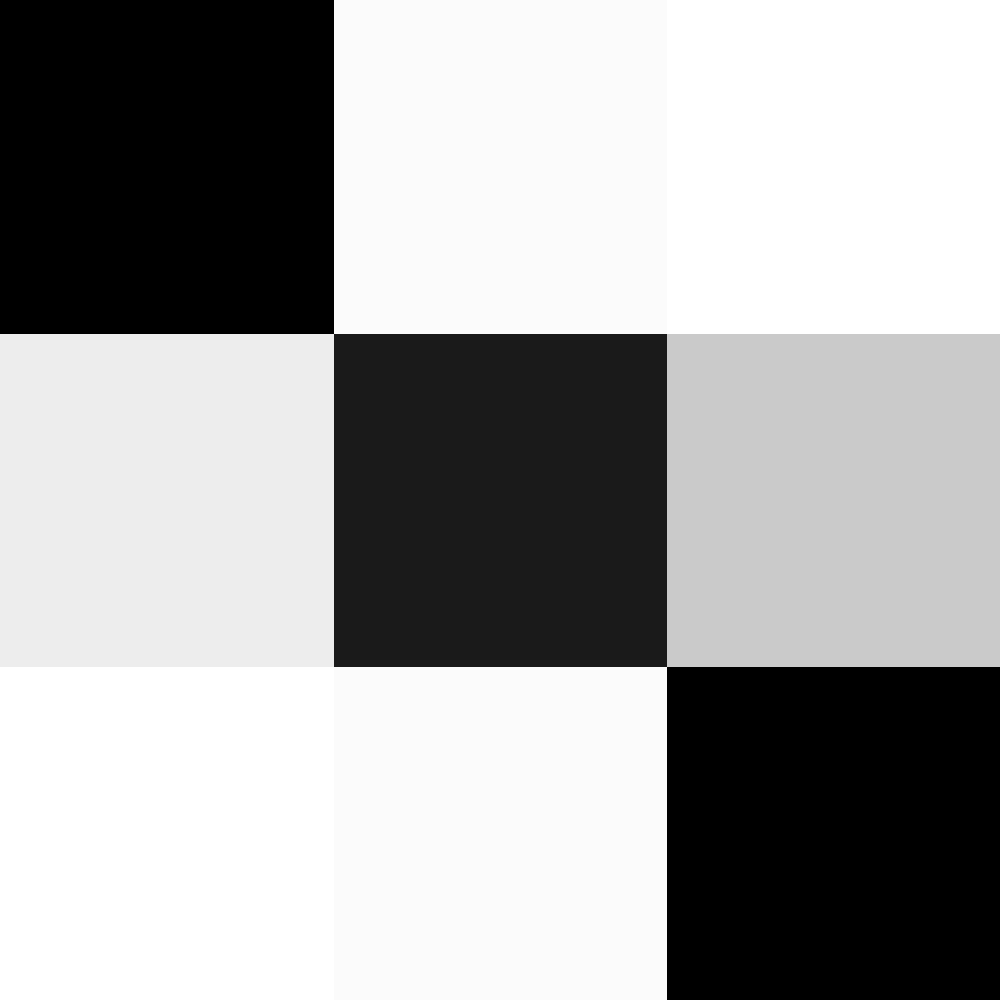} &
    \includegraphics[width=0.25\linewidth]{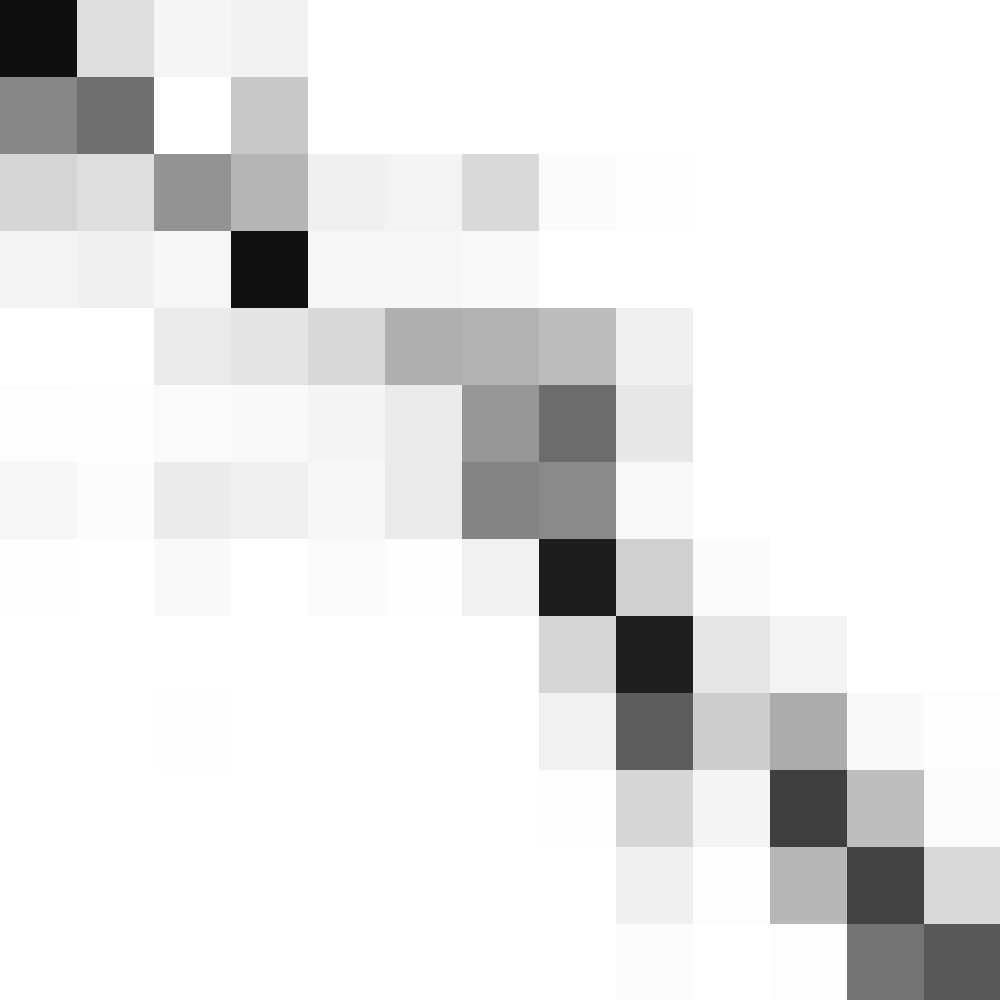} \\
    \rotatebox{90}{\hspace{12pt}difference} &
    \includegraphics[width=0.25\linewidth]{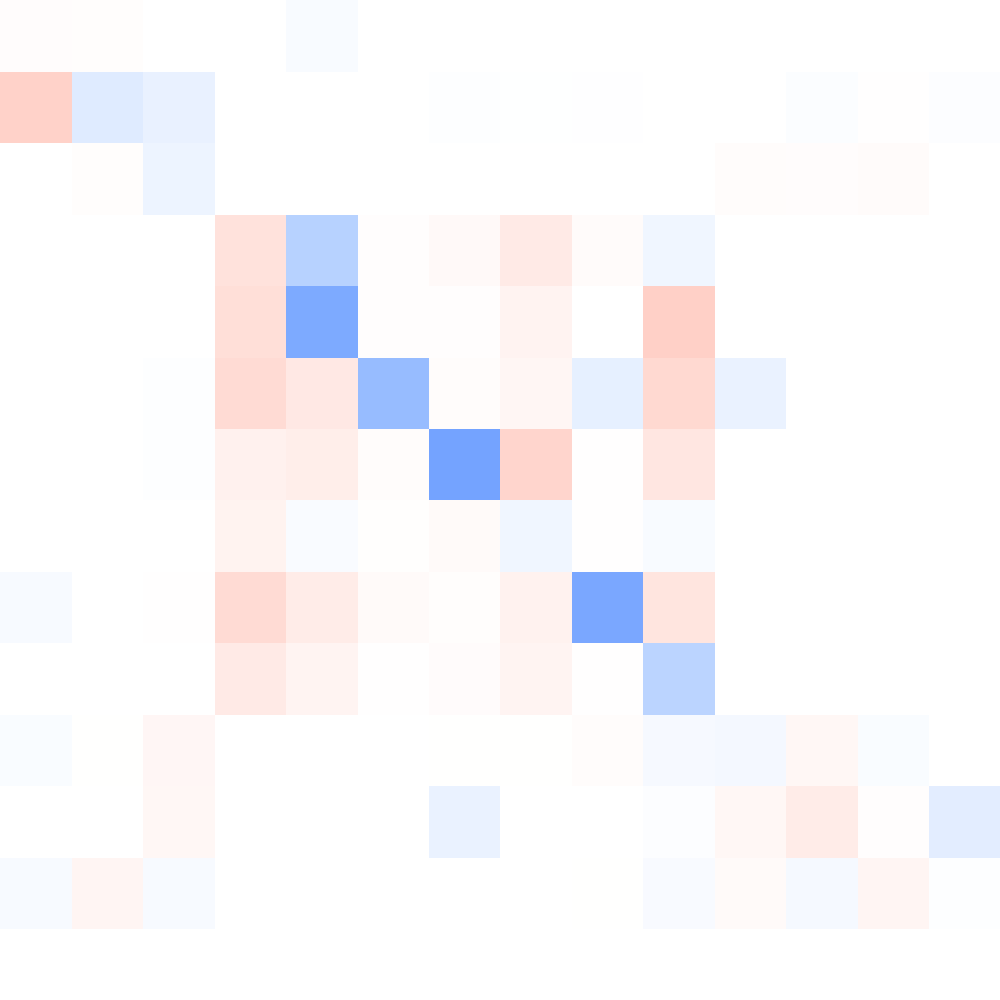} &
    \includegraphics[width=0.25\linewidth]{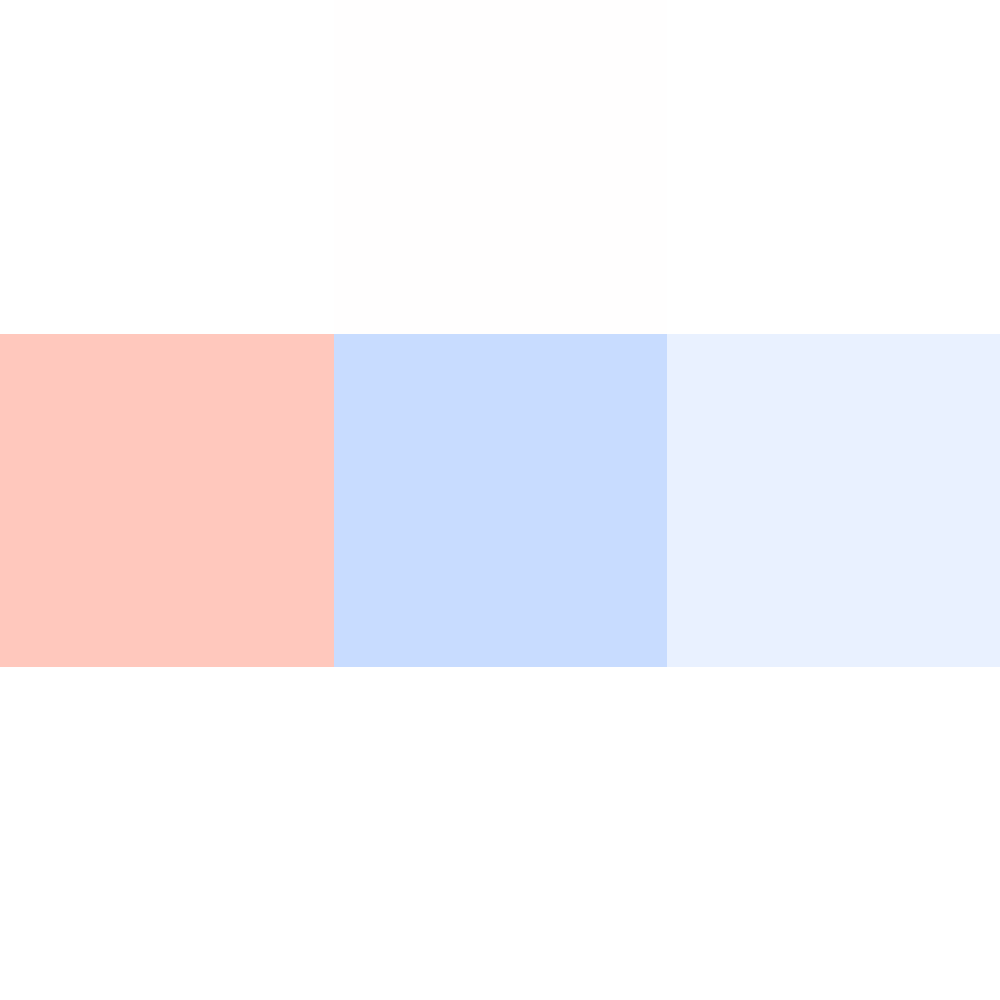} &
    \includegraphics[width=0.25\linewidth]{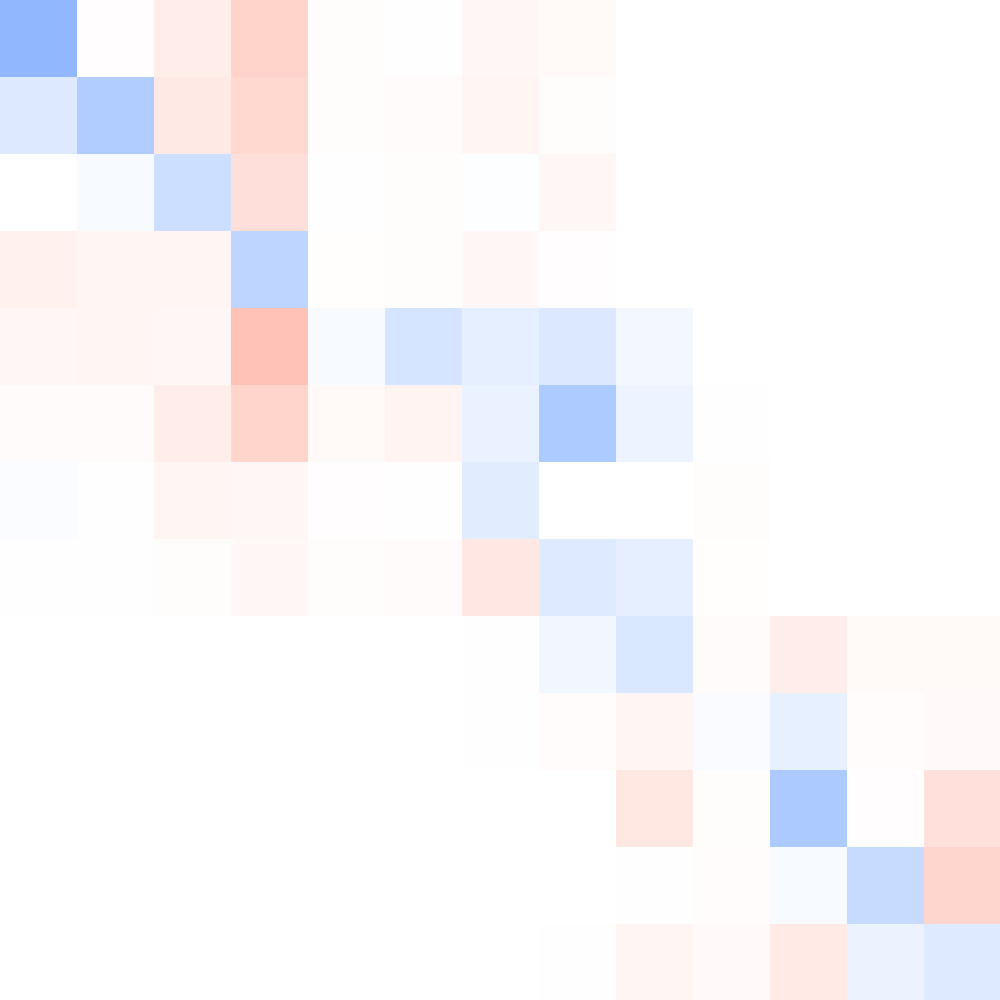} \\
    & \multicolumn{3}{c}{\footnotesize $-35\%$ \hspace{50pt} generated $-$ dataset \hspace{50pt} $+55\%$} \\
    & \multicolumn{3}{c}{\includegraphics[width=0.83\linewidth]{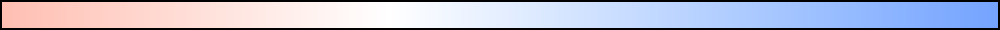}}
\end{tabular}
\caption{Classification confusion matrix on the dataset and generated samples and their difference.}
\Description{}
\label{fig:attr_mat}
\end{figure}

\paragraph{Effectiveness of Attribute Control.}
We evaluate whether, for any given label, the distribution of the features in the images produced by our method closely match the real data distribution. For this purpose, we train a classifier for age, gender and ethnicity, independent from the generator training process. The classifier is trained on 85\% of normalized data and tested on the other 15\%. We then generate a large number of samples using random codes and attribute labels, and classify these generated samples using the classifier. If the generator controls the attributes accurately, the performance of the classifier should be similar on the test data and on the generated samples. The accuracy (average of all classes) is shown in \Cref{tab:attr_acc}, and the confusion matrix is shown in \Cref{fig:attr_mat}. The performance of the classifier on the dataset and the generated samples are similar, and where they differ, the accuracy is generally higher on the generated samples.

\begin{table}
\centering
\begin{tabular}{cccc}\\\toprule
     & race & gender & age \\\midrule
     dataset & $60.09\%$ & $84.72\%$ & $32.66\%$ \\
     generated & $75.88\%$ & $91.93\%$ & $49.90\%$ \\
     difference & $+15.79\%$ & $+7.21\%$ & $+17.24\%$ \\\bottomrule
\end{tabular}
\caption{Classification accuracy on the dataset and generated samples and their difference.}
\label{tab:attr_acc}
\end{table}

\paragraph{Identity Consistency in Attribute Editing.}
A key demand for semantic editing of faces is the consistency in facial identity as the attributes of interest (age, gender and ethnicity in our case) are adjusted. Our asset generator is trained to disentangle labeled and unlabeled information and therefore designed to excel in this aspect. However, there is no direct metric proposed in the past that can conveniently evaluate this capability quantitatively: existing face embedding networks for facial identification take all semantic information into consideration, since age, gender and ethnicity are posed as part of one's identity in these models. 

We therefore devise a similarity metric based on linear fitting to evaluate identity preservation of our model in gender and age editing. With the existing face embedding network OpenFace~\cite{amos2016openface}, we synthesize a large number of random pairs of images, where the two codes for each pair are identical except the controlled attributes. Then, the difference vectors of the embeddings are computed from the two images. The large set of difference vectors are used for linear fitting, where the fitted line should represent the direction corresponding to the controlled attributes (e.g. gender, age) in the embedding space. Note that we exclude ethnicity from this evaluation since its dimensionality is unknown. 

The gender-and-age-independent identity similarity can therefore be obtained by projecting the face embeddings onto the complement of the subspace spanned by the estimated gender and age directions. The similarity of two images can be measured by the cosine similarity. We therefore compare the identity similarity of pairs of images where gender and age attributes are edited by our models against random pairs of face images. The results are shown in \Cref{tab:similarity}, where we find that our model supports attribute editing that maintains a high similarity in identity under our metric. If we change just the age and/or gender, the result should have high similarity to the original, while two random images should have lower similarity.



\begin{table}
\centering
\begin{tabular}{ccc}\\\toprule
     change gender & change age & random pair \\\midrule
     $0.9594$ & $0.9542$ & $0.8703$ \\\bottomrule
\end{tabular}
\caption{Identity Similarity between pairs of rendered images. Under our proposed directional metric, we find that our model supports gender and age editing that maintains a high similarity in identity.}
\label{tab:similarity}
\end{table}

\paragraph{Evaluation on the Two-Step Disentanglement Training.}
To evaluate the effectiveness of the disentanglement training, we compare our two-stage training design to a typical conditional GAN model as the baseline approach. Examples are shown in \Cref{fig:gan_inversion_comparison}. We find that our generator can better reproduce unlabeled features (e.g. facial identity): In column 1, the result produced by the baseline GAN has flatter eyebrows; in columns 2 and 4 the baseline result displays less facial hair; in column 3 the baseline result appears more masculine; in column 5 the baseline result shows less wrinkles.

\subsubsection{Application}
\begin{figure}
\captionsetup{type=figure}
\includegraphics[width=1\linewidth]{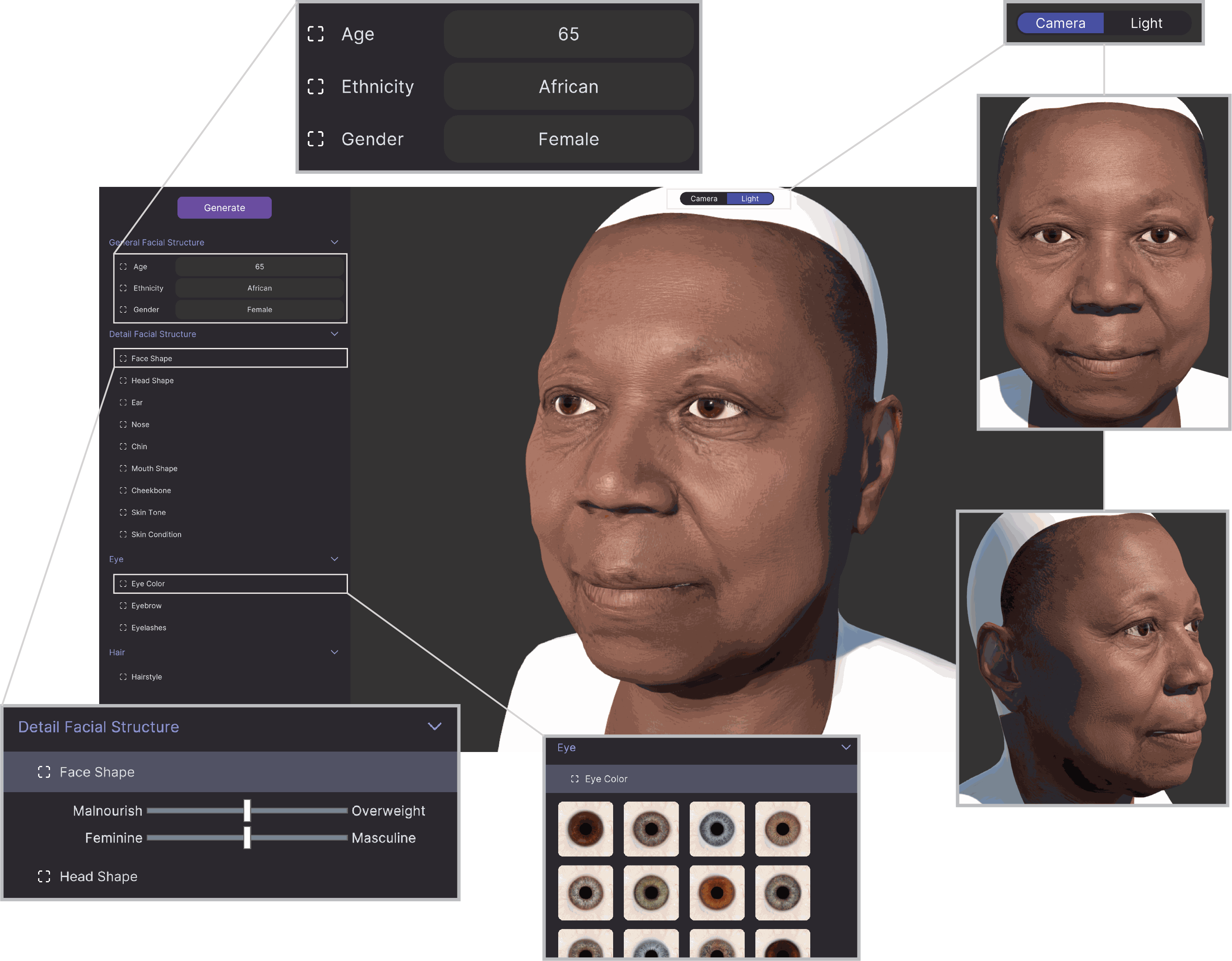}
\caption{Interface of the avatar creation web application. Users set up features to generate and refine the facial structure and features of the avatar through an interactive and intuitive graphical representation.}
    \Description{}
\label{fig:ui_supp}
\end{figure}

We design an on-the-fly interactive web application based on the proposed method to validate its downstream application value. \Cref{fig:ui_supp} visualizes the interface for a 3D face asset creation and editing tool. The users are given options to specify "General Facial Structure" attributes such as age, ethnicity, and gender, and "Detailed Facial Structure" features including the shape of the face, nose, ears, chin, and more. The web application supports free-viewport manipulation and change of lighting for detailed inspection of the generated results. The application employs a front-end system for rendering and a back-end system for providing services and hosting the web pages. The front-end system uses the Unity game engine~\cite{Unity2022} and is built against the WebGL platform. All queries on editing age, ethnicity, and gender are processed by the back-end system, which runs our proposed neural networks to generate the corresponding assets. The mesh object, albedo map, specular map, and displacement map are the major art assets dynamically generated by our models and saved on the cloud storage associated with an ID in our database for later reference. Changes to other facial feature settings are handled locally on the browser, and no new assets are generated by the server. This design eliminates significant delay to ensure an interactive editing experience. In a local session, related assets, such as eye color textures and geometry offset presets, are stored on the back-end server and only loaded on demand.
Changes to these modifiable features are updated in the database to allow reproducibility. The back-end system is implemented using ASP.NET Core~\cite{ASPNETCore} and modularized. \textit{Please refer to our supplementary materials for video recordings of the web application}.

\section{Limitations}
While we have made considerable progress towards a production-quality system for semantic-guided generation of PBR face assets, there are aspects of our system that can be improved upon.

By sourcing our training data from diffusion models, we expanded the diversity of our training data greatly. However, our method only works for texture maps, as there is currently no controllable high-quality large-scale generative model for face geometry. While we can refine the geometry with a single-view face reconstruction method, the diversity of the initial geometries is still limited by the scanned dataset. 

Our texture normalization matches diffusion-generated UV textures to the scanned-albedo domain. But its accuracy is limited by the small scanned target domain (200 subjects), which can reduce skin-tone fidelity for rare appearances. We therefore estimate a skin color from manually selected flat regions as additional input. With richer scanned-albedo coverage, leveraging general-purpose intrinsic decomposition as additional de-lighting supervision is a potential way to improve normalization further.

Furthermore, the invisible parts of the UV texture projected from the frontal view portraits, generated by the diffusion model, need to be supplemented by the scanned dataset. This requires a well-distributed scan dataset. If the generated portrait has no similar labels in the scanned data, the skin texture may not be accurately completed. Therefore, we aim to explore a multi-view consistent texture synthesis method that can improve our texture completion design.

Since pre-trained diffusion models are used to create the training data, our generation model may inherit any potential biases present in these models. Although we ensure that each attribute has balanced training data through manually defined attributes and distributions in data preparation, it is worth discussing how our method might still inherit issues if the pre-trained models do not express certain attributes diversely. Additionally, due to the high quality of the generated faces, additional care should be taken to prevent misuse and protect privacy.

\section{Conclusion}

We have introduced a novel 3D face generative model that allows for semantic control and creates high-quality face models, which is unprecedented in 3DMM of human faces. The main contributions of our model are a specifically designed normalization network and a disentangled generator, which can leverage not only high-quality scanned models but also in-the-wild face models, which are abundant in quantity and semantic information. Experiments have demonstrated that our model can effectively normalize faces under arbitrary lighting conditions, generate novel faces, and perform attribute manipulations on the generated 3D face in multiple semantic directions. We believe that the progress our system achieves has great potential for use in many applications, including VFX production, customized digital avatars, and the generation of synthetic training data for other fundamental computer vision research.

\begin{acks}
This research was sponsored by the U.S. Army Research Laboratory (ARL) under contract number W911NF-14-D-0005. The views and conclusions contained in this document are those of the authors and should not be interpreted as representing the official policies, either expressed or implied, of the Army Research Office, the U.S. Army Research Laboratory or the U.S. Government. The U.S. Government is authorized to reproduce and distribute reprints for Government purposes notwithstanding any copyright notation.
\end{acks}

\bibliographystyle{ACM-Reference-Format}
\bibliography{paper}

\clearpage
\appendix
\section{Details of the Comparisons and User Study}

Six short textual prompts, each describing a distinct subject identity in terms of age, gender, and ethnicity, were used to generate 3D human models with each method. These prompts served as the primary input to the generation process and were also shown to participants during evaluation. The full list of prompts included: ``Elderly Asian Female,'' ``Young Adult Western Female,'' ``Teenage African Male,'' ``Young Western Male,'' ``Young Asian Female,'' and ``Middle-Aged African Male.'' All other inputs and settings followed the default configurations or recommended usage described in the original implementations of DreamFace~\cite{zhang2023dreamface} and Describe3D~\cite{wu2023high}.

For each prompt, participants viewed the outputs generated by three different methods. The three outputs were displayed in randomized order. The sequence of prompts was also randomized for each participant.

Each output was rated using two questions on a five-point Likert scale (\textit{Very Poor} to \textit{Excellent}): (1) ``How well does the model match the description?'' and (2) ``How would you rate the skin detail quality?'' In a small number of trials, a third question was included as an attention check, instructing participants to select a specific response (e.g., ``Please select 'Poor'.''). Participants who failed one or more of these checks were excluded from the analysis.
\section{Hair and Accessories}

To support more complete and realistic digital head models, our system includes hair and common accessories such as eyeglasses, hats, and masks. Hair is modeled using Maya XGen and is attached to a predefined scalp mesh. Since all head models share the same topology, the hair can be transferred across different identities by deforming it according to the shared vertex correspondence.

As for accessories, eyeglasses are rigged individually, and hats and masks are deformed using lattice-based transformations, allowing them to fit varying face geometries.
\section{Challenging Cases}
\begin{figure}[htbp]
    \centering
    \includegraphics[width=\linewidth]{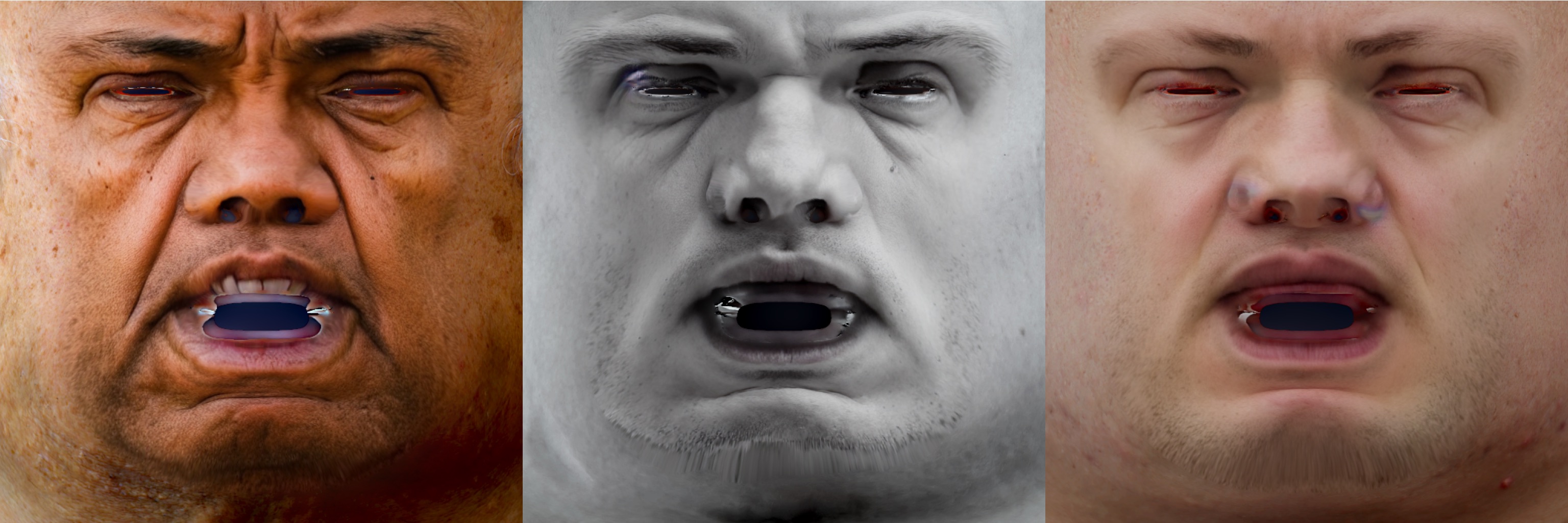}
    
    \vspace{0.5em}
    \noindent
    \parbox[b]{0.33\linewidth}{\centering (a)}%
    \parbox[b]{0.33\linewidth}{\centering (b)}%
    \parbox[b]{0.33\linewidth}{\centering (c)}
    
    \caption{
        Common artifacts in textures resulting from directly applying the proposed processing method to face images generated by diffusion models.
        (a) malformed teeth;
        (b) unnatural color tone;
        (c) distortions near the nose.
    }
    \label{fig:challenging_sd}
\end{figure}

\begin{figure}[t!]
 \centering
 \includegraphics[width=\linewidth]{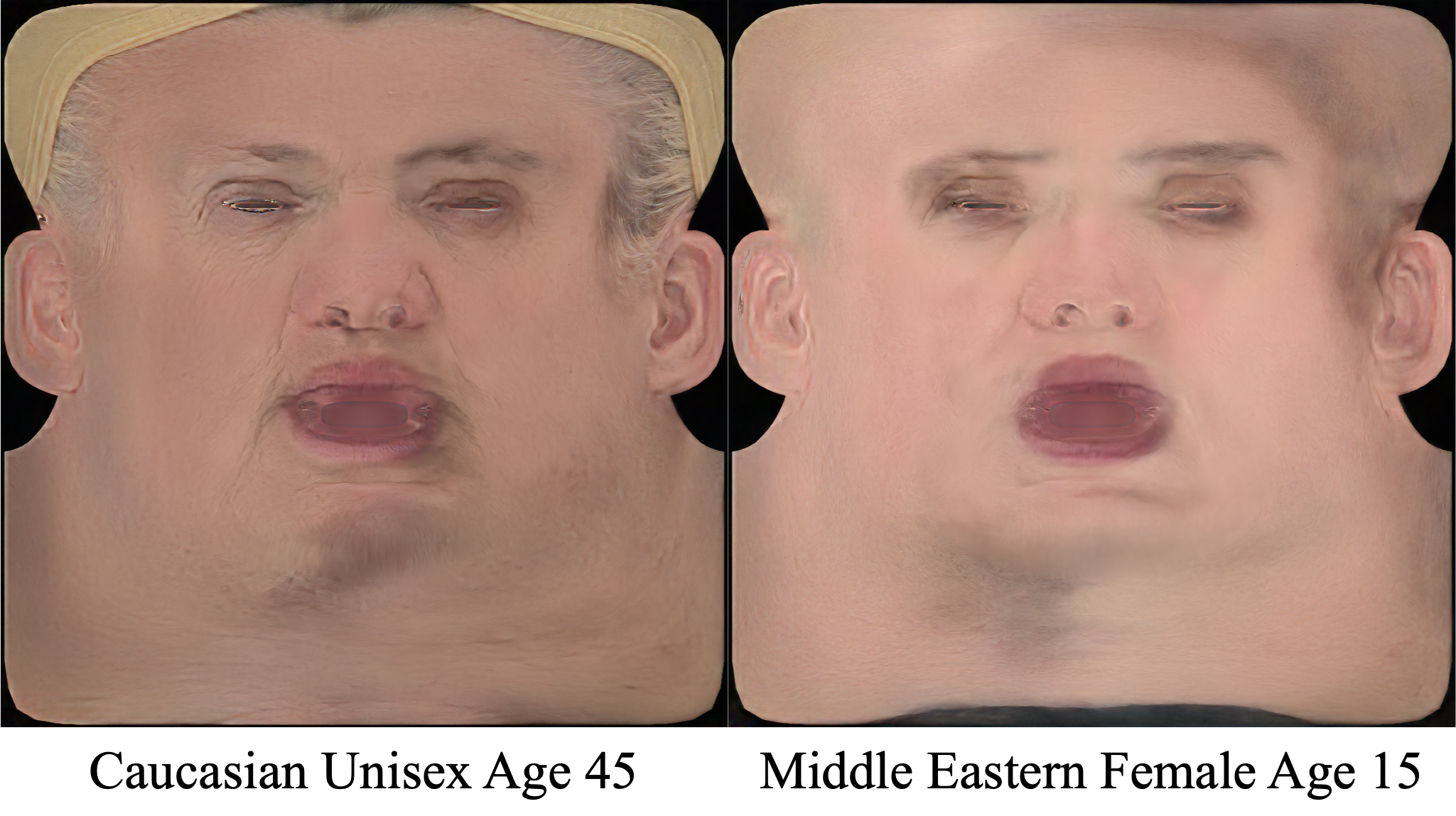}
 
\caption{Common artifacts in textures generated by proposed GANs: lighting baked in for some parts; unnatural/uneven color tone.}
\label{fig:challenging_gan}
\end{figure}

The examples in ~\Cref{fig:challenging_sd} demonstrate that directly processing images generated by general-purpose diffusion models to UV space often leads to significant artifacts, such as malformed teeth, unnatural tone, and local geometric distortions. Some are due to imperfect control over the diffusion model, where the final synthesized image does not align pixel-perfectly with the input geometry, causing mismatches in the UV projection. Others are inherent to the generative process. The model produces undesired color tones due to a lack of structural constraints. Although we manually filtered out abnormal textures when constructing 44,000 UV textures, these issues highlight the need for a generative model specifically trained for texture synthesis.
\section{Dataset demographic distribution}
\begin{figure}[t!]
 \centering
 \includegraphics[width=\linewidth]{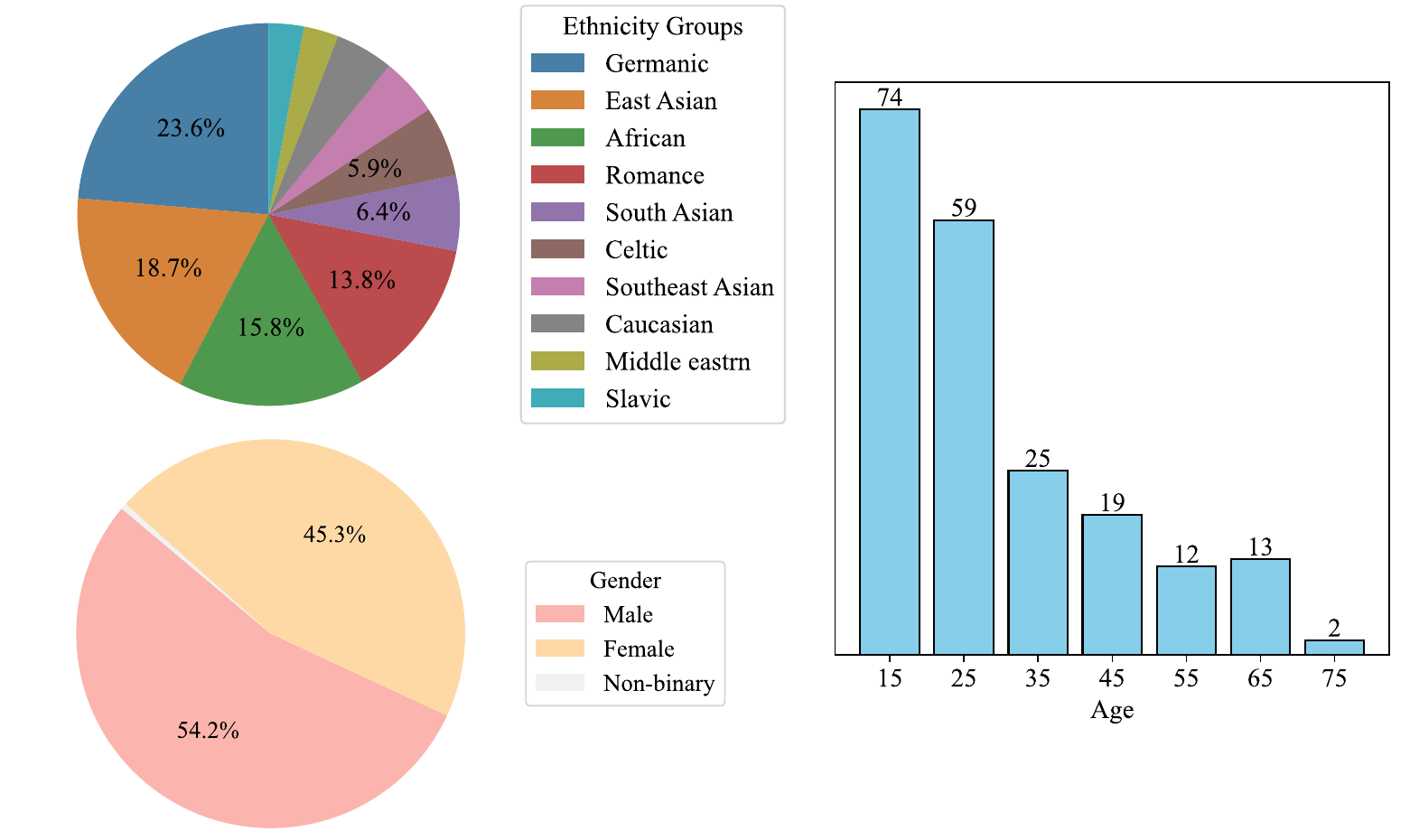}
 
\caption{Overview of the demographic distribution of the scanned database. It illustrates the distribution of ethnicity, age, and gender groups, respectively, in our high-quality dataset. These categories are used solely as generative control labels and do not constitute anthropological claims.}
\label{fig:data_dist_fig}
\end{figure}

The \Cref{fig:data_dist_fig} illustrates the distribution of demographic information in our high-quality scanned dataset.

\end{document}